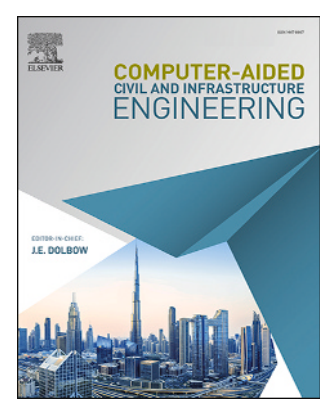

Research Article

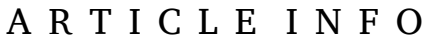

# Perception-and-action system for humanoid robot task execution in construction

Yanxi Liu [a,1], Yizhi Liu [b,1,*]

[a] Department of Civil and Environmental Engineering, Syracuse University, 158 Link Hall, Syracuse, NY 13244, USA
[b] Department of Civil and Environmental Engineering, Syracuse University, 151J Link Hall, Syracuse, NY 13244, USA

ABSTRACT

Humanoid robots, with their human-like shape and multi-tasking capabilities, are well-aligned with human-dominated workplaces, like those in civil and construction engineering, where they could collaborate with human workers or autonomously perform physically demanding and hazardous tasks. Despite this promise, limited research has explored how to endow these robots with the practical capabilities needed to perform construction tasks. To this end, this study proposes a novel perception-and-action system that enables humanoid robots to learn and perform construction tasks from worker demonstrations. This system contains two deep networks: Humanoid-PoseNet, which extracts human postures and translates them into mechanically feasible poses for a humanoid robot; and Humanoid-ActionNet, which learns robot-executable actions based on these translated poses. Experimental results demonstrate that the humanoid robot reliably executed eight construction-related actions, achieving an average motion-tracking error of 82.45 mm MPJPE (Mean Per Joint Position Error). This work provides an early step toward deploying humanoid collaborators in construction.

## 1. Introduction

Construction sites rank among the most complex and dynamic work environments, requiring labor-intensive tasks and posing significant safety risks (Hu et al., 2025). According to the International Labor Organization (International Labour Organization, 2015), construction accidents account for roughly 30% of all work-related deaths worldwide, underscoring the hazardous nature of these sites. Furthermore, a combination of rising infrastructure demands and a shortage of skilled labor (Deloitte, 2026) has intensified the call for automation. The U.S. Bureau of Labor Statistics (U.S. Bureau of Labor Statistics, 2025) projects that overall employment in construction and extraction occupations will grow faster than the average for all occupations from 2024 to 2034, with about 649,300 openings projected each year on average due to employment growth and worker replacement needs, creating a pressing need for innovative solutions to maintain productivity and ensure worker well-being. As a result, automation and robotics have become increasingly vital in meeting both the demand for new infrastructure and the imperative of workplace safety (Li et al., 2025; Liu et al., 2026; Vähä et al., 2013).

To this end, the construction industry has seen growing adoption of a series of robots that collaborate with workers to perform specific tasks in the field (Ojha et al., 2026). For instance, the Semi-Automated Mason (SAM) robot has been deployed to the construction field to collaborate with workers for bricklaying tasks (Soto et al., 2020); the MuLE robot has been designed to assist human workers to lift heavy materials (Construction Robotics, 2018); and the Spot robot (legged robot), has been deployed to work alongside workers for site inspection and monitoring (Wetzel et al., 2022). These robots have demonstrated promising efficiency gains by improving precision, reducing labor-intensive workloads, and enhancing workplace safety. In parallel with these robot deployments, studies – including the authors' previous work – have explored a series of mechanisms to facilitate safe and effective interaction between workers and these robots on construction tasks (Liang et al., 2020; Liu et al., 2025, 2021; Liu & Jebelli, 2024). Such mechanisms include enabling robots to interpret workers' states and avoid collisions, thereby promoting smoother, more coordinated human-robot interaction.

Beyond these construction robots already in use, there is another type of robot that also shows significant potential for serving as an assistive partner to collaborate with workers on construction tasks: humanoid robots (Sheng et al., 2025; Tong et al., 2024). Compared to deployed construction robots, humanoid robots offer distinct advantages, making them another promising robotic solution in construction. First, construction sites are human-dominated (Zhang et al., 2021) – tools, scaffolding, stairs, ladders, and workspaces are all built to human dimensions. Therefore, the human-like shape of the humanoid robots can facilitate more natural collaboration with onsite personnel. Second, the dexterity and high degree of freedom (DoF) of the humanoid robots (Özbaltan et al., 2025) allow them to distinguish from the widely-adopted collaborative construction robots that can mainly perform a specific task (i.e., single-task construction robots), which have the potential of working with human collaborators to perform a series of tasks. For example, a single humanoid robot could deliver materials, assist with assembly, help workers perform inspections, and handle finishing tasks (Dao et al., 2024) – reducing the need for multiple single-task collaborative robots.

* Corresponding author.
*E-mail addresses:* yliu557@syr.edu (Y. Liu), yliu580@syr.edu (Y. Liu).
[1] These authors contributed equally to this work

Despite these advantages, deploying humanoid robots on construction sites remains challenging, as evidenced by the absence of any humanoid robots currently in operation in the field. Several issues contribute to this absence: (a) construction tasks require specialized domain knowledge (Prieto et al., 2024): for each type of task, workers may need special behaviors or actions and tools to perform, such knowledge is not inherently available to robots; (b) the varying task execution demands, uneven working conditions, and contact-rich movements present a great challenge to maintaining stability and balance of the humanoid robot (Gu et al., 2025) – particularly for its bipedal locomotion system. Given these challenges, direct deployment of a humanoid robot on-site will lead to inefficiency and safety concerns for both robot and human workers. These issues raise a central research question in this study: How can humanoid robots learn and execute construction-related tasks efficiently and safely?

Inspired by the broadly similar body structures of humanoid robots and human workers, a natural direction for addressing this question is to enable humanoid robots to learn construction-related actions directly from how workers execute tasks. Workers' motions and actions provide observable expressions of construction task execution, such as task-specific whole-body coordination and tool- and material-handling behaviors. Therefore, encoding these motions and actions can provide a useful source of task-execution information for humanoid robot learning (He et al., 2024b).

Human pose provides a structured representation for encoding workers' motions and actions and converting them into machine-learnable data for humanoid robots, as it can quantitatively capture body-joint locations, limb relationships, and temporal kinematic patterns across task execution (Fan et al., 2024; Zhu et al., 2023). However, learning from worker poses is not straightforward for humanoid robots. Two key technical gaps need to be addressed. First, although existing human pose-estimation methods (Liu & Jebelli, 2024; Pavlakos et al., 2018) can recover human postures from visual observations, their outputs are generally human-centered. Due to differences in limb proportions, joint structures, range of motion, and actuation constraints, these human poses cannot be directly used as humanoid-compatible motion references for task learning. Second, even after human motion is translated into humanoid-compatible poses, a gap remains in enabling the humanoid robot to convert these pose references into physically executable whole-body actions while maintaining balance, contact consistency, and motion stability in practical construction environments (He et al., 2025).

To bridge these gaps and position the humanoid robot as a reliable robotic collaborator on construction sites, this study proposed a Vision-based Perception-and-Action (VPA) system composed of two modules: Humanoid-PoseNet and Humanoid-ActionNet. Humanoid-PoseNet enables the humanoid robot to perceive worker demonstrations and retarget the observed human motions into humanoid-compatible pose trajectories. Humanoid-ActionNet learns a whole-body control policy that executes retargeted humanoid poses while maintaining the robot's dynamic stability. The authors evaluate the proposed VPA system through a case study in which a Unitree humanoid robot learns and performs representative construction-related actions (e.g., carrying a pipe, holding bricks, and lifting boxes) in both simulation and lab environments. To the best of the authors' knowledge, this study provides an early step toward enabling humanoid robots to learn and execute construction tasks from human demonstrations. Moreover, this study is expected to facilitate the implementation of robotics in the construction industry.

The remainder of this paper is structured as follows: Section 2 provides a literature review related to the proposed VPA system, covering current developments in human pose estimation and whole-body control of humanoid robots. It also discusses the remaining challenges in applying these techniques to humanoid construction task learning and summarizes the technical contributions of the proposed study to address these challenges. Then, Section 3 details the proposed VPA system for humanoid robots, including the architecture and training procedure of the Humanoid-PoseNet and Humanoid-ActionNet. In Section 4, the designed case study will be introduced to evaluate the proposed methodology. Accordingly, the evaluation results will be reported in Section 5. Section 6 further discusses the evaluation results, comparative findings, and technical implications of this study. Finally, Section 7 concludes the study by summarizing the main contributions, limitations, and future research directions.

## 2. Related work

### 2.1. Existing studies supporting the development of the VPA system for humanoid construction task execution

As introduced, despite the potential alignment between humanoid robots and construction-site demands, there is still no established research that enables humanoid robots to learn and execute real-world construction tasks. To this end, the authors propose the VPA system that builds on two core research domains: (i) (vision-based) human pose estimation and (ii) whole-body control of humanoid robots. Pose-estimation methods provide a basis for capturing workers' task-relevant motion information, enabling humanoid robots to obtain task-execution representations, such as kinematic patterns and limb coordination, from workers' task executions. Whole-body control methods are then needed to enable humanoid robots to learn from these representations and convert them into executable humanoid actions for construction task execution. Existing studies in these two domains provide important foundations, but gaps remain when these methods are applied to humanoid task learning and execution. The following review summarizes representative studies in these domains and highlights the challenges that motivate the design choices of this study.

Researchers in fields like computer science, human dynamics, and civil engineering have developed a series of vision-based pose estimation techniques that can allow humanoid robots to learn workers' posture, including two-dimensional (2D) human pose estimation and 3D pose estimation. 2D human pose estimation aims to predict the pixel coordinates of key body joints to generate a body skeleton in 2D space. Methods like Convolutional Pose Machine (Wei et al., 2016) and OpenPose (Cao et al., 2021) were representative methods that can extract human keypoints from image frames, which have been implemented in construction contexts. For example, Xiong and Tang applied an OpenPose-based method to detect construction workers' 2D skeletons from 2D images (Xiong & Tang, 2021). Moreover, the authors' prior works developed CPM-based 2D pose estimation methods (Liu et al., 2024; Liu & Jebelli, 2022) to estimate construction workers' posture and ergonomic risks. However, 2D pose estimation is limited by its lack of depth: multiple distinct 3D human body configurations can project to the same 2D keypoints. This ambiguity of 2D pose estimation limits a humanoid robot's ability to accurately learn and reproduce actions that require precise 3D spatial structure and motion. 3D human pose estimation addresses this limitation by estimating human postures in 3D space. Existing approaches produce two types of outputs: (i) 3D keypoint-based poses and (ii) parametric body-model-based poses. Keypoint-based methods infer the 3D coordinates of human body joints from images or videos, which is essential for transferring human motion to a humanoid robot that requires full 3D kinematics to reproduce posture. Much recent work focuses on monocular 3D pose estimation, where a single camera feed and learned models predict 3D joint positions. Notable works in computer vision demonstrated that even a simple neural network can lift 2D keypoints to 3D with reliable accuracy (Martinez et al., 2017), and that incorporating temporal information from video (e.g., VideoPose3D by Pavllo et al., 2019) further smooths and improves 3D predictions. The authors' prior work, 3D-PoseNet (Liu & Jebelli, 2024), similarly enabled a robotic system (UR5e robotic arm) to perceive workers' posture in 3D during bricklaying tasks. Beyond keypoints, parametric body-model approaches estimate 3D posture by fitting a structured human body model (Bogo et al., 2016; Kolotouros et al., 2019; Loper et al., 2015). A leading example is SMPL (Skinned Multi-Person Linear model), a learned model that represents 3D human pose and shape with continuous parameters and outputs a deformable 3D mesh (Kanazawa et al., 2018; Kocabas et al., 2020). In other words, compared with keypoints, SMPL provides a richer surface-level representation of the human body. For example, Pavlakos et al. proposed an end-to-end ConvNet framework to estimate SMPL parameters from a single RGB image (Pavlakos et al., 2018). Subsequent work extended this formulation to SMPL-X by integrating articulated hands and expressive facial components within a unified model (Pavlakos et al., 2019). In construction, Chu et al. used SMPL-based estimation to assess worker ergonomic risks (Chu et al., 2019), and Zhang et al. proposed Construction worker Voxel Action Recognition network (CVAR-net) to recognize worker actions from SMPL-based posture representations (Zhang et al., 2025). In practice, these two families also involve a trade-off: keypoint-based methods are more lightweight, while SMPL-based methods provide more anatomically consistent reconstructions at a higher computational cost.

While these methods can enable humanoid robots to capture human posture from the visual system, a critical gap remains for humanoid learning: morphological mismatch between the human body and the humanoid robot (He, Luo et al., 2024; McCrory et al., 2025). Differences in limb proportions, joint structure, and range of motion mean that an accurate 3D human pose cannot be directly understood by a humanoid robot, limiting its usefulness for humanoid action learning.

Moreover, even if the human-to-humanoid morphological mismatch is addressed, pose estimation typically captures only "surface-level" kinematics (i.e., how the body moves) and does not, by itself, model physical contact and interaction with the environment – such as maintaining contact stability and preserving whole-body balance during contact-rich behaviors (Barreiros et al., 2025; Belvedere et al., 2024; Ferrari et al., 2023). Enabling these capabilities requires whole-body control (WBC), defined as the humanoid's ability to coordinate movements across the arms, legs, torso, and base to physically interact with the environment and perform real-world tasks (Chen et al., 2025). While achieving efficient WBC remains a challenge due to the high degrees of freedom (DoF) and nonlinear dynamics of humanoid systems (Chen et al., 2025; Ze et al., 2025), researchers have made promising progress in recent years (Cheng et al., 2024b; Lu et al., 2025; Myers et al., 2025). Lu et al. introduced Mobile-TeleVision, which uses a VR headset to control upper-body motions and foot pedals to command lower-body locomotion (Lu et al., 2025). Similarly, Myers et al. introduced an IMU-based interface that enables joint-level whole-body control through wearable inertial sensors (Myers et al., 2025). While effective for demonstration, teleoperation does not produce autonomous task execution and typically requires operators to wear additional hardware – an approach that may be impractical and potentially unsafe in dynamic construction environments (e.g., a VR headset can block the operator's field of view and reduce situational awareness on site) (Cheng et al., 2024b). To enable humanoids to control their motions while maintaining stability and balance without continuous human supervision, researchers have explored reinforcement learning (RL)-based whole-body control (Cheng et al., 2024; Yang et al., 2025). For example, Cheng et al. proposed Expressive Whole-Body Control (ExBody), which uses RL policies to generate coordinated full-body behaviors such as walking, dancing, and gesturing (Cheng et al., 2024). Radosavovic et al. similarly trained a humanoid controller using model-free RL in simulation and demonstrated efficient real-world locomotion across varied terrain (Radosavovic et al., 2024). Despite these advances, current RL-based controllers remain difficult to deploy for construction tasks because they are developed and evaluated in daily-life settings; construction-specific actions and contact conditions are rarely considered in the current RL-based controller design.

### 2.2. Contributions

Motivated by the limitations discussed in Section 2.1, this study proposes a Vision-based Perception-and-Action (VPA) system that enables humanoid robots to learn construction skills directly from video-recorded worker demonstrations. The proposed system (i) infers humanoid-compatible 3D keypoints from worker demonstrations, and (ii) learns a control policy that allows the humanoid to execute the learned behaviors for construction tasks. The main technical contributions are summarized as follows.

- First, the authors design Humanoid-PoseNet, a vision-based perception module for a humanoid robot to learn from human demonstrations. Unlike conventional human-pose outputs from pose-estimation methods, the representations generated by Humanoid-PoseNet are designed to support human-to-humanoid learning by accounting for the mismatch between worker body motion and the humanoid robot's kinematic structure. This module provides the pose-level bridge between visual worker demonstrations and downstream humanoid action learning.
- Second, the authors develop Humanoid-ActionNet, a teacher-student RL architecture that learns humanoid kinematics from the humanoid representations extracted by Humanoid-PoseNet. Compared with existing RL-based WBC methods, which are mainly designed for general human-like behaviors, Humanoid-ActionNet enables the humanoid robot to physically execute construction-related actions while maintaining dynamic feasibility, including stability, balance, and contact consistency. This design supports more reliable sim-to-real transfer in construction settings.

Moreover, to the best of the authors' knowledge, this is the first method that enables a humanoid robot to learn and execute construction tasks from human demonstrations. The following section will detail the proposed methodology.

## 3. Methodology

### 3.1. Overview

Fig. 1 illustrates an overview of the proposed VPA system for humanoid robots, which integrates Humanoid-PoseNet and Humanoid-ActionNet. As shown, Humanoid-PoseNet includes a 2D-to-3D pose estimation network component and a human-to-humanoid retargeting network component. The pose estimation network reconstructs 3D human keypoints from 2D video frames captured during worker demonstrations. Given the reconstructed 3D human pose, the human-to-humanoid retargeting network maps the human motion into the humanoid's configuration space via a decoder-latent space-encoder structure, generating retargeted humanoid pose trajectories. Notably, these retargeted trajectories describe only the kinematic (surface-level) motion from video without information about contacts, forces, or physical feasibility in the real world. Therefore, Humanoid-ActionNet feeds these retargeted trajectories into a teacher-student reinforcement learning (RL) framework with physics-aware rewards to learn pose-aware actions that remain dynamically stable under realistic environmental interactions. In this study, the rewards are described as physics-aware because they are designed based on the physical execution requirements of the humanoid robot. The reward terms encourage joint-level motion tracking, keypoint tracking, root velocity tracking, body orientation control, upper-body coherence, foot-contact consistency, and foot-slip reduction. These terms guide the humanoid robot to follow the retargeted demonstration while maintaining balance, contact stability, and smooth whole-body coordination. To be more specific, the teacher policy is trained in simulation to maximize task performance while enforcing physical feasibility (e.g., balance and contact consistency). The student policy then distills the teacher's behaviors into a representation with reduced reliance on full-body proprioception (joint positions/velocities, root kinematics), promoting motion stability and contact integrity under real-world environments and enabling reliable sim-to-real transfer. Overall, by integrating Humanoid-PoseNet and Humanoid-ActionNet, the humanoid robot can learn posture-driven construction skills directly from worker demonstrations. The following sections describe each part in detail. In addition, the proposed method will be evaluated in Section 4.

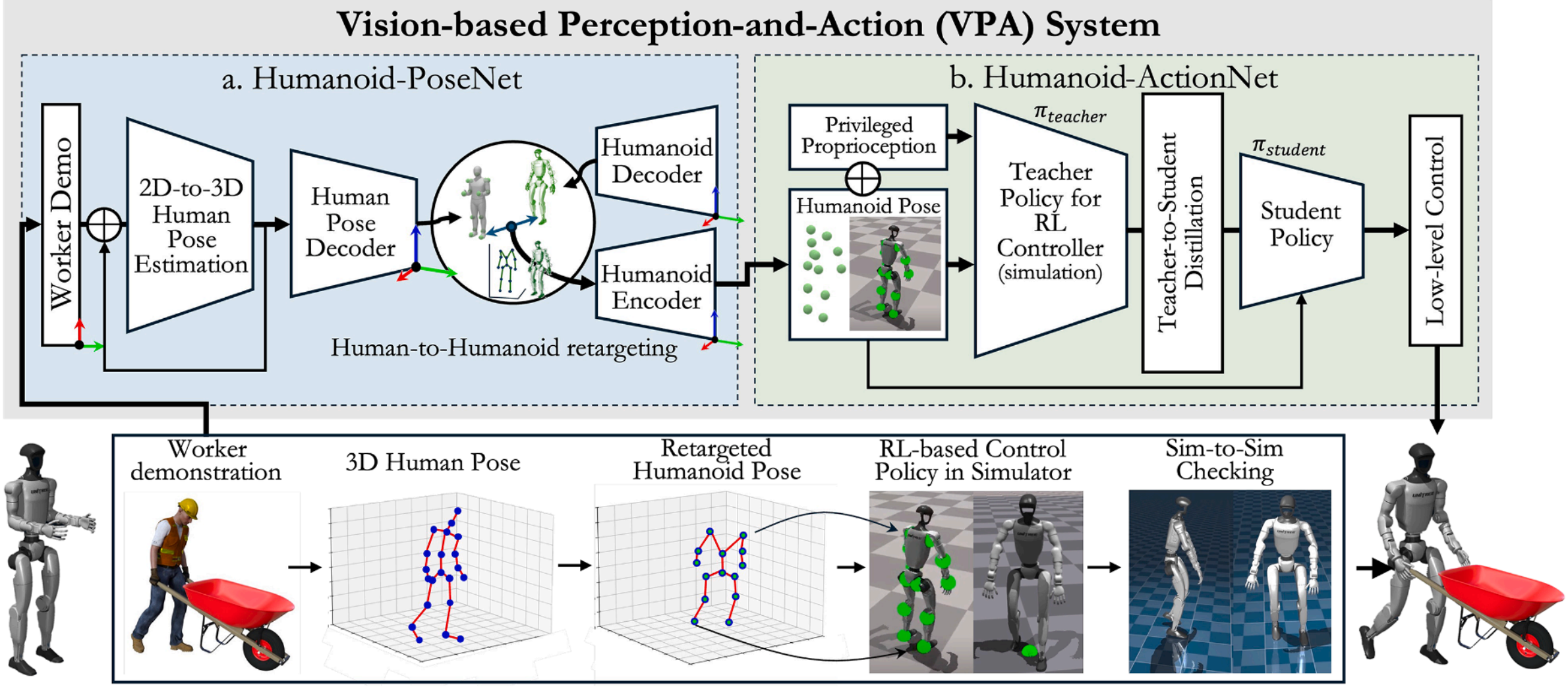

**Fig. 1.** Overview of the proposed VPA system for humanoid construction task execution.

### 3.2. Humanoid-PoseNet – vision-based deep network for estimating 3D humanoid posture from human demonstrations

As shown in Fig. 2, Humanoid-PoseNet consists of two parts: (1) a 3D human pose estimation module that reconstructs workers' 3D posture from 2D RGB images/videos, and (2) a human-to-humanoid motion retargeting network that maps the extracted 3D human pose into the humanoid's 3D configuration. As reviewed in Section 2.1, the vision-based 3D human pose estimation method is a mature research area, and several off-the-shelf models can meet the accuracy and robustness requirements of this study. Accordingly, the authors evaluated representative models – including PoseNet (Moon et al., 2019), VideoPose3D (Pavllo et al., 2019), AlphaPose (Fang et al., 2022), and 3D-PoseNet (Liu & Jebelli, 2024) – for extracting 3D human sequences from RGB inputs, denoted as $\mathscr{P}^{3D}_{h,i}$ $(i = 1, \ldots, N)$. The results show that these methods achieve comparably reliable 3D pose reconstruction performance and can serve as suitable "upstream" pose-estimation models for the subsequent components of the proposed VPA system. In addition, since 3D pose estimation is not the primary contribution of this study, architectural details of these models will not be introduced. Sections 5 and 6 provide the quantitative comparisons and discussion.

Regarding the second part of Humanoid-PoseNet, the human-to-humanoid retargeting module, Fig. 2 shows its details: it consists of two encoders ($E_h$ and $E_r$) and one decoder ($D_r$) with a shared latent space (i.e., white circle in Fig. 2). Specifically, the human pose encoder $E_h$ maps an estimated 3D human pose $\mathscr{P}^{3D}_{h,i}(i \in \{1,\dots,N\})$ into the latent space as $z_i \in \mathbb{R}^d$, while the robot pose encoder $E_r$ maps a humanoid pose $\mathscr{P}^{3D}_{r,k}$ into the same latent space ($z_k \in \mathbb{R}^d$). Both $E_h$ and $E_r$ are implemented as Multi-Layer Perceptron (MLP) networks with eight fully connected (FC) hidden layers, each with 128 hidden units and ReLU activations. Their output layer produces a $d$-dimensional latent feature. The decoder $D_r$ maps each latent embedding (e.g., $z_i$) back to the humanoid pose space, expressed as $\widehat{\mathscr{P}}^{3D}_{r,i} = D_r(z_i)$. The decoder mirrors the encoders design with eight FC hidden layers (each has 128 units and ReLU), and its output layer is designed to generate a pose vector in the humanoid's configuration space with dimension $d = 14$ in this study.

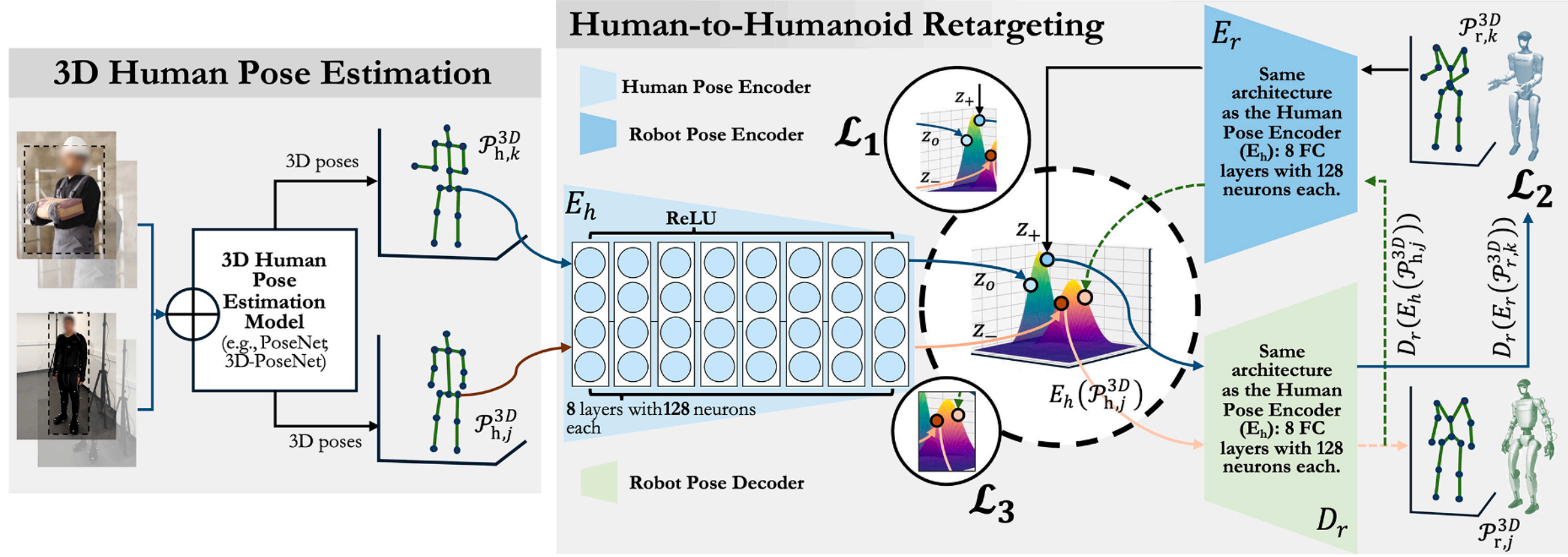

**Fig. 2.** Overall structure of the Humanoid-PoseNet.

During the training, encoders in this network receive pose triplets sampled from the human and robot pose dataset (refer to Section 4 for more details): for each pose triplet, $E_h$ takes two human poses $\mathscr{P}^{3D}_{h,\,i}$ and $\mathscr{P}^{3D}_{h,j}$ $(i \neq j)$; and $E_r$ takes one humanoid robot pose $\mathscr{P}^{3D}_{r,k}$. Then, these three are used to produce latent embeddings ($z_i,\ z_j,\ z_k$). The latent space randomly selects one of ($z_i,\ z_j$) as the anchor $z_o$ (e.g., $z_o = z_i$). Next, a pose-aware similarity score is computed in pose space between the anchor-associated pose and the remaining human and robot poses – $\mathscr{P}^{3D}_{h,\,i}$ vs. $\mathscr{P}^{3D}_{h,\,j}$ and $\mathscr{P}^{3D}_{h,\,i}$ vs. $\mathscr{P}^{3D}_{r,\,k}$ – and use the similarity scores to assign the corresponding embeddings as the positive (more similar to the anchor) or negative (less similar) samples in latent space. To achieve this, the network measures this similarity using the bone-direction angular error (BAE) (Liu et al., 2022):

$$BAE = \frac{1}{|B|}\sum_{(a,b)\in B}\cos^{-1}\left(clip\left(u^h_{ab}\cdot u^r_{ab},-1,1\right)\right) \quad (1)$$

Where $B$ is the set of matched bones (links) between the human and humanoid skeletons, $u^h_{ab}$ and $u^r_{ab}$ are the unit bone direction vectors for the corresponding limb ($a \rightarrow b$) computed from human pose $\mathscr{P}^{3D}_h$ and humanoid pose $\mathscr{P}^{3D}_r$, respectively. The $clip(\cdot)$ indicates the clip operator that is used to avoid numerical instability when the dot product slightly exceeds $[-1,1]$. Using Equation.1, the sample with smaller BAE relative to the anchor is labeled as the positive $z_+$, and the sample with a larger BAE is labeled as the negative $z_-$. These two values are then used to formulate a triplet objective loss in latent space (with a constant $c = 0.3$ in this study):

$$\mathscr{L}_1 = \max(\ \|z_o - z_+\|_2 - \|z_o - z_-\|_2 + c, 0) \quad (2)$$

The design of this loss is inspired by contrastive learning methods reported in Radford et al. (2021). During training, this loss pulls low-BAE poses together and pushes high-BAE poses apart. This structure reduces cross-domain mismatch by encouraging human poses and their humanoid counterparts to map to nearby latent embeddings, improving retargeting stability. It also enhances generalization, because unseen human poses that are geometrically similar to previously observed motions are more likely to fall into latent regions that decode to feasible humanoid configurations (Annabi et al., 2024).

In addition to this triplet loss, the model calculates two more loss functions to train the encoders and decoder to improve the efficiency of human-to-humanoid retargeting. As shown in Fig. 2, the decoder $D_r$ maps latent embeddings $z_-$ ($z_j$) and $z_+$ ($z_k$) to humanoid pose space, producing (i) a retargeted humanoid pose decoded from a human embedding $\widehat{\mathscr{P}}^{3D}_{r,j} = D_r(z_j) = D_r\left(E_h\left(\mathscr{P}^{3D}_{h,j}\right)\right)$ and (ii) a reconstructed humanoid pose decoded from a robot embedding: $\widehat{\mathscr{P}}^{3D}_{r,k} = D_r(z_k) = D_r\left(E_r\left(\mathscr{P}^{3D}_{r,k}\right)\right)$. Because $z_+$ ($z_k$) is generated by encoding the robot pose data $\mathscr{P}^{3D}_{r,\,k}$, a standard reconstruction loss (Equation.3) is used to ensure that $D_r$ can reconstruct humanoid configurations from robot-latent inputs by minimizing the discrepancy between $\widehat{\mathscr{P}}^{3D}_{r,k}$ and $\mathscr{P}^{3D}_{r,\,k}$. In addition, decoding a human embedding $z_j = E_h\left(\mathscr{P}^{3D}_{h,j}\right)$ may generate a pose that drifts away from the robot's feasible configuration space. To reduce this issue, a latent-consistency loss (Equation.4) is introduced. This loss encourages the decoded pose $\widehat{\mathscr{P}}^{3D}_{r,j}$ to be encoded by $E_r$ into a latent representation consistent with the original human embedding, thereby promoting cross-domain distribution alignment during retargeting, as demonstrated in Aberman et al. (2020); Yan et al. (2024). The two loss functions are defined as follows:

$$\mathscr{L}_2 = \|\mathscr{P}^{3D}_{r,k} - D_r\left(E_r\left(\mathscr{P}^{3D}_{r,k}\right)\right)\|_2 \quad (3)$$

$$\mathscr{L}_3 = \|\ E_h\left(\mathscr{P}^{3D}_{h,j}\right) - E_r\left(D_r\left(E_h\left(\mathscr{P}^{3D}_{h,j}\right)\right)\right)\|_2 \quad (4)$$

The overall objective is the weighted sum of Equation.2 to Equation.4, where the weights $\lambda_1 = 9.3$ and $\lambda_2 = 4.7$ tuned in this study:

$$\mathscr{L}_{\text{total}} = \lambda_1\mathscr{L}_1 + \lambda_2\mathscr{L}_2 + \mathscr{L}_3 \quad (5)$$

Finally, the authors train the network by minimizing $\mathscr{L}_{\text{total}}$ using the gradient descent. After convergence, given an estimated human pose $\mathscr{P}^{3D}_{h,t}$, the humanoid target pose is obtained by $\widehat{\mathscr{P}}^{3D}_{r,t} = D_r\left(E_h\left(\mathscr{P}^{3D}_{h,t}\right)\right)$, which serves as the retargeted humanoid configuration for the following robot control task. Section 4 provides details on the human pose and humanoid pose datasets used in this study to train and evaluate the proposed network.

### 3.3. *Humanoid-ActionNet – teacher-student reinforcement learning framework for whole-body humanoid task execution*

Given the humanoid pose trajectories from Humanoid-PoseNet, $\widehat{\mathscr{P}}^{3D}_{r,t}$ ($t = 1,\dots,T$), the authors develop a whole-body controller that enables the humanoid robot to not only imitate surface-level postures (i.e., $\widehat{\mathscr{P}}^{3D}_{r,t}$), but also physically execute these behaviors under realistic dynamics and contact interactions. As shown in Fig. 3, the method adopts a teacher-student architecture consisting of a teacher layer and a student layer. In the teacher layer, a teacher policy $\pi_{\text{teacher}}$ is trained by using an observation set that comprehensively captures the information required for physically feasible task execution with action $a_t$ at time step $t$. The teacher layer receives three groups of inputs: (a) privileged observations $s^{\text{po}}_t$, including the Degree of Freedom (DoF)/joint difference between the target motion and the robot's actual motion, keypoint difference, and root velocity; (b) proprioceptive states $s^{\text{ps}}_t$, including DoF positions, DoF velocities, the previous action, root angular velocity, and root orientation (roll, pitch, and yaw); and (c) motion tracking targets $s^{\text{mt}}_t$, including the target DoF positions, keypoint positions, root linear/angular velocity, roll, pitch, yaw, and height). All three input groups are available in simulation (IsaacGym in this study) and can be directly obtained from the humanoid pose generated in Section 3.2. The selection of these inputs follows existing humanoid RL studies (Cheng et al., 2024; He, Luo, He et al., 2024; Ji et al., 2025), which demonstrate that combining privileged signals, proprioceptive feedback, and motion-tracking targets provides sufficient information to train policies that achieve efficient task execution in simulation and transfer reliably to physical humanoids. In the student layer, the teacher policy ($\pi_{\text{teacher}}$) is distilled into a deployable student policy ($\pi_{\text{student}}$) that can control the robot to perform the same tasks using only information that is available on the physical robot. $\pi_{\text{student}}$ receives the same motion tracking targets $s^m_t$, but replaces privilege observation (i.e., $s^{\text{po}}_t$) with a window of past proprioceptive states ($s^{\text{ps}}_{t-w+1},\dots s^{\text{ps}}_t$; window size $w$ is 10 in this study) that summarizes recent proprioceptive history. This teacher-student design improves robustness in realistic settings where privileged states ($s^{\text{po}}_t$) are unavailable, and it reduces the sim-to-real gap by ensuring the deployed policy depends only on information that can be obtained on the physical humanoid (He, Luo, He et al., 2024; Ji et al., 2025).

Specifically, $\pi_{\text{teacher}}$ is trained using a Proximal Policy Optimization

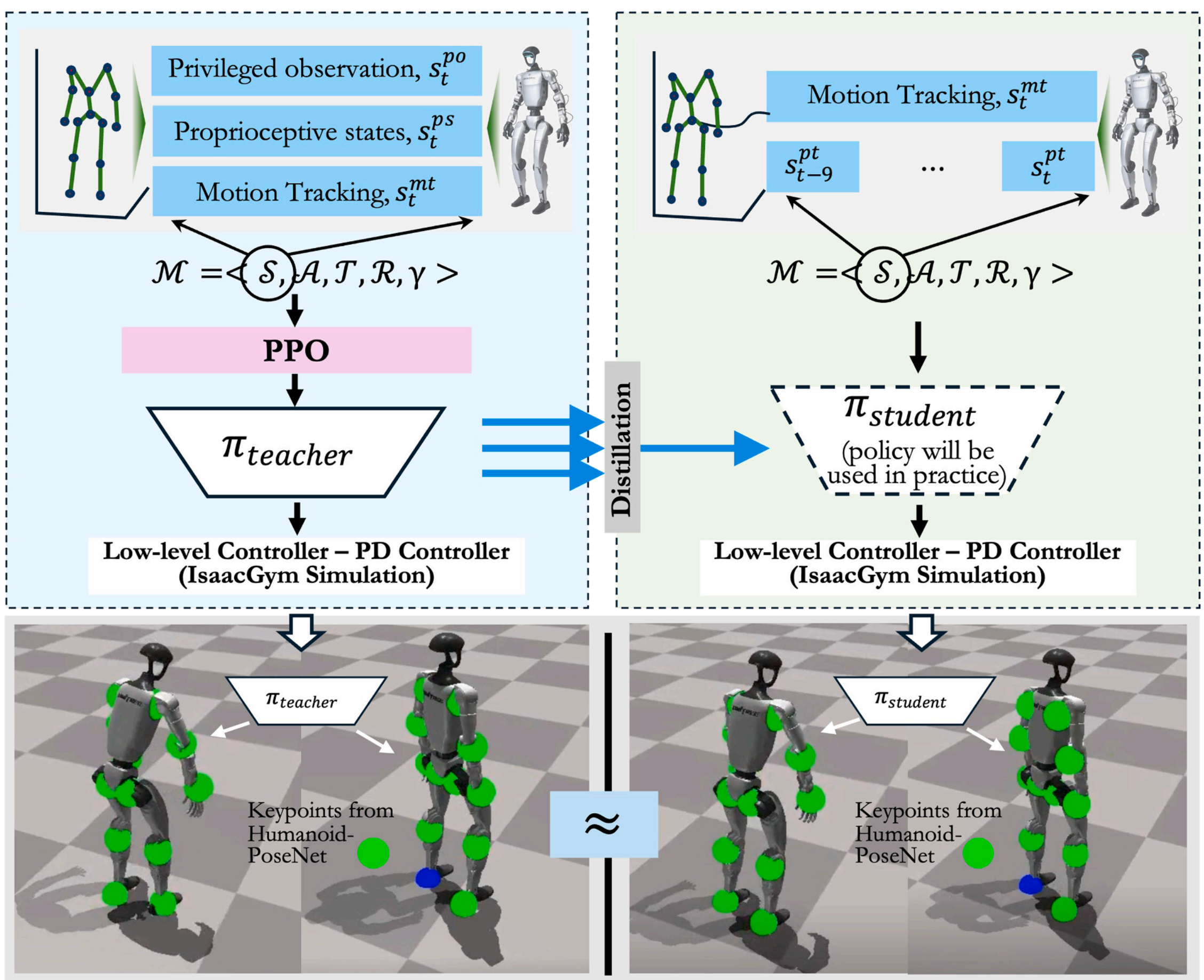

**Fig. 3.** Overall structure of the proposed Humanoid-ActionNet with teacher-student reinforcement learning architecture and PPO optimization techniques.

(PPO)-based actor-critic framework (Schulman et al., 2017). To formalize PPO training for $\pi_{\text{teacher}}$, the humanoid control is modeled as a Markov Decision Process (MDP), $\mathscr{M} \leq \mathscr{S}, \mathscr{A}, \mathscr{T}, \mathscr{R}, \gamma >$. The state $s_t = \{s_t^{\text{po}}, s_t^{\text{ps}}, s_t^{\text{mt}}\} \in \mathscr{S}$ contains the privileged observation $s_t^{\text{po}}$, proprioceptive state $s_t^{\text{ps}}$ and the motion tracking targets $s_t^{\text{mt}}$; at each time step, the policy outputs an action $a_t \in \mathscr{A}$ consisting of target joint angles for the humanoid's actuated joints. Given $(s_t, a_t)$, the MDP transitions to the next state $s_{t+1} \sim \mathscr{T}(\cdot|s_t, a_t)$, and returns a reward $r_t = \mathscr{R}(s_t, a_t) = \sum_i w_i \cdot r_i(t)$. The reward is designed to encourage the humanoid to reproduce worker-demonstrated behaviors with dynamic feasibility, and its components are summarized in Table 1. Upon $\mathscr{S}$, $\mathscr{A}, \mathscr{T}$, and $\mathscr{R}$, the objective is to learn an optimal policy $\pi^*$ that maximizes the expected reward return with a discount factor $0 < \gamma < 1$:

$$\pi^* = \underset{\pi_\theta}{\text{argmax}} \mathbb{E}_{\pi_\theta} \left[ \sum_{t=1}^{T} \gamma^{t-1} r_t \right] \tag{6}$$

PPO is then used to optimize Equation.6. PPO provides a balance between training stability and practical performance for humanoid robot control in this study. Unlike value-based methods (e.g., Q-learning and DQN-based methods (Mnih et al., 2015), which require searching over actions and are not suited to continuous action spaces, PPO can optimize a stochastic policy within an actor-critic structure. In addition, PPO's clipped surrogate objective prevents overly large policy shifts to ensure the stability of robot control, as proved by several legged and humanoid locomotion studies (Peng et al., 2018; Song et al., 2024). Fig. 4 shows the structure of the implemented PPO: the actor network (policy network in PPO) parameterized by $\theta$ represents $\pi_\theta(a_t|s_t)$ and is responsible for generating the humanoid action. In parallel, the critic network (value network in PPO) parameterized by $\phi$ learns a value function $V_\phi(s_t)$ that predicts the expected discounted reward return from state $s_t$. These two networks collaborate to optimize the control policy via the rollout and actor-critic update loop (Duan et al., 2016). At each iteration, the policy from the previous iteration $\pi_\theta^{\text{old}}$ is frozen and used to collect humanoid trajectories: for each time step $t$, the actor samples an action $a_t \sim \pi_\theta^{\text{old}}(\cdot|s_t)$ given state $s_t$, the environment calculates a reward $r_t$ based on Table 1 and returns the next state $s_{t+1}$; the critic network provides a bootstrap estimate of the current state $V_\phi(s_t)$. The action log-probability $log\pi_\theta^{\text{old}}(a_t|s_t)$ is also recorded. All quantities $\{s_{t+1}, s_t, a_t, r_t, V_\phi(s_t), log\pi_\theta^{\text{old}}(a_t|s_t)\}_{t=1,\dots,T}$ are stored in the rollout buffer.

Next, the critic's value estimates $V_\phi(\cdot)$ are used to compute the advantage estimate $\widehat{A}_t$ using generalized advantage estimation (GAE) (Schulman et al., 2015), which measures how much better the actor's action performed compared to the critic's expectation:

$$\delta_t = r_t + \gamma(1 - d_t)V_\phi(s_{t+1}) - V_\phi(s_t); \ t = 1, .., T \tag{7}$$

$$\widehat{A}_t = \delta_t + \gamma\lambda(1 - d_t)\widehat{A}_{t+1}; \ \widehat{A}_{T+1} = 0 \tag{8}$$

Where $r_t$ is the reward; $d_t \in \{0, 1\}$ is "done" indicator stored in the rollout buffer, which $d_t = 1$ indicates stop at time $t$, and $d_t = 0$ indicates a continuing transition; $\gamma$ is the discount factor, which is 0.998 in this study; and $\lambda = 0.95$ is the GAE decay parameter. This $\widehat{A}_t$ will be normalized to zero mean and unit variance.

At the same time, the actor network calculates a Gaussian policy over joint-angle targets (i.e., $a_t$):

$$\pi_\theta(a_t|s_t) = \mathscr{N}(\mu_\theta(s_t), \ diag(\sigma^2)) \tag{9}$$

$\mathscr{N}(\cdot)$ expresses the multivariate Gaussian distribution; $\mu_\theta(s_t)$ is the

**Table 1**
Reward functions designed for teacher-student RL policy.

| Reward component | Expression of reward component $r_i(t)$ | Weight $w_i$ |
|---|---|---|
| DoF position tracking | $\exp\left(-0.5\lvert q_{\text{ref}} - q\rvert\right)$ | 5.0 |
| Keypoint position tracking | $\exp\left(-0.5\lvert p_{\text{ref}} - p\rvert\right)$ | 3.9 |
| Root linear velocity | $\exp\left(-4\lvert v_{\text{ref}} - v\rvert\right)$ | 6.0 |
| Root velocity direction | $\exp\left(-4\cos\left(v_{\text{ref}}, v\right)\right)$ | 6.0 |
| Body rotation tracking | $\exp\left(-0.1 \parallel \theta_{\text{ref}} \ominus \theta \parallel\right)$ | 20.0 |
| Upper-body coherence | $\exp\left(-0.5 \parallel p_{\text{ref}}^{\text{upper}} - p^{\text{upper}} \parallel + \parallel R_{\text{ref}}^{\text{torso}} - R^{\text{torso}} \parallel\right)$ | 2.0 |
| Feet air-time shaping | $T_{\text{air}} - 0.33$ | −0.81 |
| Feet slip penalty | $\parallel v^{\text{feet}} \parallel^2 \cdot 1\left(F_{\text{feet}} \geq 1\right)$ | −0.002 |
| Regularization rewards | Refer to (Cheng et al., 2024; He, Luo, He et al., 2024; Ji et al., 2025) | As in Cheng et al. (2024); He, Luo, He et al. (2024); Ji et al. (2025) |

***Notations:** $q$ is the measured DoF positions (joint angles) from simulator; $p$ denotes the measured keypoint positions from simulator; $v$ is the measured root linear velocity; $\theta$ is the root orientation (roll-pitch-yaw); $p^{\text{upper}}$ denotes the measured upper-body keypoint positions; $R^{\text{torso}}$ is the torso orientation from the robot state; $T_{\text{air}}$ is the elapsed time since last ground contact for the swing foot; $1\left(F_{\text{feet}} \geq 1\right)$ is a binary foot-contact indicator that equals 1 when the contact force exceeds 1 N; $v^{\text{feet}}$ denotes the foot tangential velocity. Moreover, $q_{\text{ref}}$, $p_{\text{ref}}$, $v_{\text{ref}}$, $\theta_{\text{ref}}$, $p_{\text{ref}}^{\text{upper}}$, and $R_{\text{ref}}^{\text{torso}}$ are corresponding reference (target) quantities, which can be obtained from the retargeted poses by Humanoid-PoseNet and/or the ground-truth data in Section 4.

action-mean vector generated by the actor network; and $diag(\sigma^2)$ denotes the diagonal covariance matrix whose elements are the learned per-DoF action variances. During training, the executed action $a_t$ is sampled from this Gaussian: $a_t = \mu_\theta(s_t) + \sigma \odot \mathcal{N}(0, I)$, $\odot$ is the element-wise multiplication.

Given $\pi_\theta(a_t|s_t)$ and the stored $\pi_\theta^{\text{old}}(a_t|s_t)$, PPO updates the actor and critic at every time step $t$ by first computing the probability ratio:

$$\rho_t(\theta) = \frac{\pi_\theta(a_t|s_t)}{\pi_\theta^{\text{old}}(a_t|s_t)} = \exp\left(log\pi_\theta(a_t|s_t) - log\pi_\theta^{\text{old}}(a_t|s_t)\right) \tag{10}$$

This ratio is used to calculate the loss function of the actor network:

$$\mathcal{L}_{\text{actor}}(\theta) = \mathbb{E}_t[\max(-\rho_t(\theta)\widehat{A}_t,\ -clip(\rho_t(\theta),\ 1-\epsilon,\ 1+\epsilon)\widehat{A}_t)] \tag{11}$$

Here, the function $clip(\cdot)$ clamps the ratio into $[1-\epsilon,\ 1+\epsilon]$ (with $\epsilon =$ 0.2 in this study) to avoid overly large policy changes and stabilize the training process. Relying on this loss, the policy is updated via the gradient descent process $\theta \leftarrow \theta - \alpha\nabla_\theta\mathcal{L}_{\text{actor}}(\theta)$ over mini-batches sampled from the rollout buffer.

In parallel, the critic is trained using the empirical return targets $\widehat{R}_t$:

$$\widehat{R}_t = \widehat{A}_t + V_\phi(s_t) \tag{12}$$

With the value regression loss function (Schulman et al., 2017):

$$\mathcal{L}_{\text{critic}}(\phi) = \mathbb{E}_t\left[\left(V_\phi(s_t) - \widehat{R}_t\right)^2\right] \tag{13}$$

Similarly, the critic network $\phi$ is updated via the gradient descent process over mini-batches sampled from the rollout buffer. The combined loss of the PPO process is expressed as:

$$\min_{\theta,\phi} \mathcal{L}_{\text{actor}}(\theta) + c_v\mathcal{L}_{\text{critic}}(\phi) \tag{14}$$

$c_v$ weights the value terms. This update is repeated iteratively: for each iteration, the prior iteration policy collects on-policy rollouts, and the actor-critic parameters are updated over $n$ epochs until convergence.

After optimizing the teacher policy ($\pi_{\text{teacher}}^*$), this policy is distilled to a student policy using the loss function:

$$l_{\text{distill}} = \parallel a_t^s - a_t^* \parallel^2 \tag{15}$$

Here, $a_t^s$ is the action generated by the student policy $a_t^s \sim \pi_{\text{student}}\left(s_t^{\text{student}}\right)$ given the state $s_t^{\text{student}} = \left\{\left(s_{t-w+1}^{\text{ps}}, \ldots, s_t^{\text{ps}}\right), s_t^{\text{mt}}\right\}$, $s_t^{mt}$ is still the motion tracking target generated by Humanoid-PoseNet and $(s_{t-w+1}^{\text{ps}}, \ldots, s_t^{\text{ps}})$ is a sequence of past proprioceptive states with a time window ($w$ = 10). In addition, $a_t^*$ is the action generated by the optimized teacher policy $a_t^* \sim \pi_{\text{teacher}}^*\left(s_t^{\text{teacher}}\right)$ given the state $s_t^{\text{teacher}} = \left\{s_t^{\text{ps}}, s_t^{\text{mt}}\right\}$, which $s_t^{\text{ps}}$ is the proprioceptive state and $s_t^{\text{mt}}$ is also the motion tracking target from Humanoid-PoseNet. Notably, the authors combine Equation.15 with DAgger (Dataset Aggregation) technique (Ross et al., 2011) to distill the student policy until $l_{\text{distill}}$ converges. More details of the DAgger process can be found in Ross et al. (2011).

After distillation, the deployed student policy outputs are tracked by a Proportional-Derivative (PD) controller to ensure robust and smooth execution on the real humanoid robot:

$$\tau_t = K_p(a_t - q_t) - K_d\dot{q}_t \tag{16}$$

Where $\tau_t$ is the commanded joint torque, $q_t$ and $\dot{q}_t$ are current joint position and velocity read directly from the robot, and the gains $K_p$ and $K_d$ are tuned to balance the robot stability, improving the sim-to-real transfer. Notably, in this study, the tested robotic platform is the Unitree G1, which has 23 actuated DoFs (e.g., pitch/roll of the arm) mapped from the humanoid pose representation (14 joints) in Section 3.1. Therefore, $\mu_\theta(s_t)$, $\sigma$, $a_t$, $q_t$, $K_p$ and $K_d$ are all 23-dimensional vectors ($\in \mathbb{R}^{23}$). When deploying to other humanoids (e.g., Unitree H1 or platforms with different actuation layouts), the action dimensionality and the actor's final output layer and Gaussian covariance dimension will be adjusted.

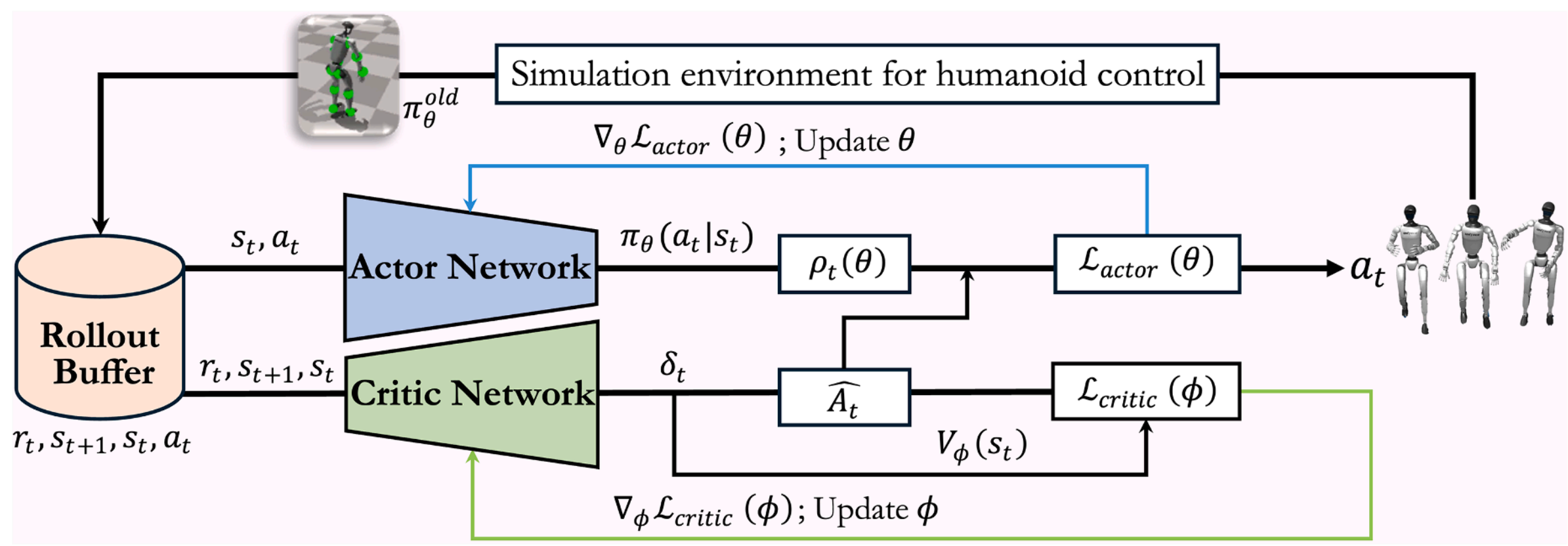

**Fig. 4.** PPO training loop for the teacher policy in Humanoid-ActionNet.

Finally, the authors report the designed structure of the actor and critic network used in PPO in this study. The actor maps the input states into the action $\mu_\theta(s_t)$ through four stages: (1) the network begins with a short-history feature encoder that summarizes recent motion trends over the short window. The motion trends include four consecutive previous frames of proprioceptive states $s_{t-3}^{\text{ps}},\ldots,\ s_t^{\text{ps}} \in \mathbb{R}^{d_{\text{ps}}}$. Each $d_{\text{ps}}$-dimensional frame was passed through a two-layer fully connected (FC) encoder consisting of an FC layer of 128 neurons followed by an Exponential Linear Unit (ELU) activation, and a second FC layer of 16 neurons followed by ELU. The resulting four embeddings (each $\mathbb{R}^{16}$) were then concatenated to form a 64-dimensional vector and fused by an additional FC layer of 16 neurons followed by ELU. This module outputs a compact feature representation $z_t^{\text{feat}} \in \mathbb{R}^{16}$; (2) the actor then appends $z_t^{\text{feat}}$ with current proprioceptive state $s_t^{\text{ps}} \in \mathbb{R}^{d_{\text{ps}}}$ and the motion-tracking features $s_t^{\text{mt}} \in \mathbb{R}^{d_{\text{ms}}}$ to preserve fine-grained current-state information; (3) a latent $z_t \in \mathbb{R}^{d_z}$ is injected: for the teacher policy, a privileged observation $s_t^{\text{po}} \in \mathbb{R}^{d_{\text{po}}}$ was extracted from the input and passed through a privileged encoder designed as a multilayer perceptron (MLP). The MLP has two fully connected layers with 64 neurons in the first hidden layer and $d_z$ neurons in the second layer. The output dimension of this pathway is denoted by $d_z$. For the student policy, $s_t^{\text{po}}$ was replaced by a proprioceptive window $\left(s_{t-9}^{\text{ps}},\ldots,s_t^{\text{ps}}\right)$, which is fed to the history encoder: this encoder began with a FC projection from 10 $\times\, d_{\text{ps}}$ to 30 neurons followed by ELU. This sequence was then processed by two 1D convolution layers (Conv1D): the first Conv1D used 20 output channels with a kernel size of 4 and a stride of 2, followed by ELU, and the second Conv1D used 10 output channels with a kernel size of 2 and a stride of 1, followed by ELU. After flattening, an FC layer mapped the resulting 30-D feature to $d_z$ neurons followed by ELU, producing the student latent $z_t$; and (4) all features are concatenated as $x_t = \left[z_t^{\text{feat}}, s_t^{\text{ps}},\ s_t^{\text{mt}},\ s_t^{\text{po}},\ z_t\right]$, and passed through a three-layer MLP, where each hidden layer contained 256 neurons followed by ELU activation. A final FC layer projected this representation to 23 neurons, yielding $\mu_\theta(s_t)$.

The critic network $V_\phi(s_t)$ is designed as a feedforward value-regression subnetwork that maps the critic observation vector (denoted as $s_t$ in implementation) to a single scalar value estimate. The network begins with a fully connected (FC) layer that takes $s_t$ inputs and expands them to 256 hidden units, followed by an Exponential Linear Unit (ELU) activation function to introduce nonlinearity (Clevert et al., 2016). After this initial projection, two additional FC layers are stacked, each maintaining 256 hidden units and followed by ELU activations. These layers refine the latent representation of the current state, enabling the critic to capture the nonlinear relationships between the humanoid's observation variables (e.g., proprioception, goal-related features, and/or privileged signals when available). Finally, the refined 256-dimensional feature vector is fed into an output FC layer with a single neuron, producing the scalar value estimate $V_\phi(s_t)$.

## 4. Case study – humanoid learning and execution of construction tasks

To evaluate the feasibility of the VPA system, the authors conducted a construction-related humanoid task-execution case study with a training session and a testing session. In the training session, the authors collected datasets to train Humanoid-PoseNet for 3D worker pose estimation and human-to-humanoid pose retargeting. Using the resulting retargeted pose trajectories, the authors then trained Humanoid-ActionNet (RL controller) in simulation to learn humanoid control policies. In the testing session, the full VPA system (Humanoid-PoseNet + Humanoid-ActionNet) was evaluated under sim-to-sim and sim-to-real transfer environments. Overall, the training session focused on developing the two learning-based components of the system, while the testing session examined whether these components could be integrated to support construction-related humanoid task execution. Details are provided below.

### 4.1. Training session – experimental setup

The authors first established a construction-oriented action-capture environment in which human subjects performed a set of construction-related behaviors. As shown in Fig. 5-(a), the experiment was conducted in an indoor workspace (2.8 $m$ in width, 5 $m$ in length, and 2.5 $m$ in height) instrumented with an OptiTrack motion-capture system comprising six infrared cameras, providing sufficient space for whole-body locomotion and manipulation demonstrations. The six tripod-mounted cameras were distributed around the capture volume to minimize demonstration occlusion (e.g., construction materials may occlude the human body) through overlapping fields of view. Each camera was mounted at around 2.3 $m$ height, and its orientation and tilt angle were adjusted relative to the tallest participant (1.87 $m$ in this study) to ensure complete body coverage across diverse postures.

Before data collection, each subject was asked to wear a lightweight, tight-fitting motion-capture suit (Fig. 5). To account for different body sizes while maintaining a close fit, the authors prepared suits in multiple sizes (L-XL) and selected the appropriate size for each subject to minimize marker motion relative to the body. Then, a standard wand-sweep and L-frame calibration were performed to verify the spatial accuracy of the motion capture system and remove blind spots. Subjects were outfitted with 41 reflective markers placed at standardized anatomical landmarks (e.g., head, shoulders, spine, hips, elbows, wrists, knees, and ankles) following the OptiTrack skeletal model (Fig. 5-(a)).

Next, each subject was asked to perform 30 construction-related actions, as reported in Table 2. Five subjects were recruited; all were healthy adults aged 19 – 25 with no reported physical or learning disabilities. Participants were civil engineering majors with prior exposure to construction practice. In addition, to further improve consistency across demonstrations, an example reference posture for each action was displayed on a large screen within the subject's line of sight (Fig. 5-(b)). Subjects performed each action at a self-selected comfortable pace. For load-bearing actions, payloads were capped at $\leq 2$ kg for lifting/carrying and $\leq 10$ kg for total cart contents, with lighter loads permitted upon request. Each action was repeated three times, with short rest intervals between repetitions to mitigate fatigue. All procedures followed approved IRB requirements. Fig. 5-(c) to -(e) provide several examples of the captured actions. Moreover, during data collection, each subject was required to complete all 30 actions in a single session (~30 min of motion recording plus ~ 5 min of rest). Each recording session began with an 8-second test capture to verify clean marker trajectories and real-time synchronization between the motion-capture and video streams. Notably, the synchronized RGB videos were paired with the corresponding 3D pose trajectories to fine-tune the 3D pose estimation component of Humanoid-PoseNet.

### 4.2. Training session – dataset preparation and model training

Using the setup in Section 4.1, a synchronized motion-capture plus RGB video dataset was collected. As introduced, five subjects performed 30 construction-related actions, and each action was repeated three times, resulting in 450 (= 5 × 30 × 3) trials. OptiTrack recorded marker trajectories and solved skeletal motion at 120 Hz, while an RGB stream recorded demonstrations at 30 fps. Across all five subjects, the dataset contains approximately 150 minutes (= 5 subjects × 30 min/subject) of demonstrations, corresponding to around 270,000 images (= 150 min × 60 s/min × 30 fps). To reduce significant pose redundancy, the authors subsampled the RGB stream by selecting one frame every 30, yielding 9000 paired RGB-3D samples for fine-tuning the 3D pose estimation module.

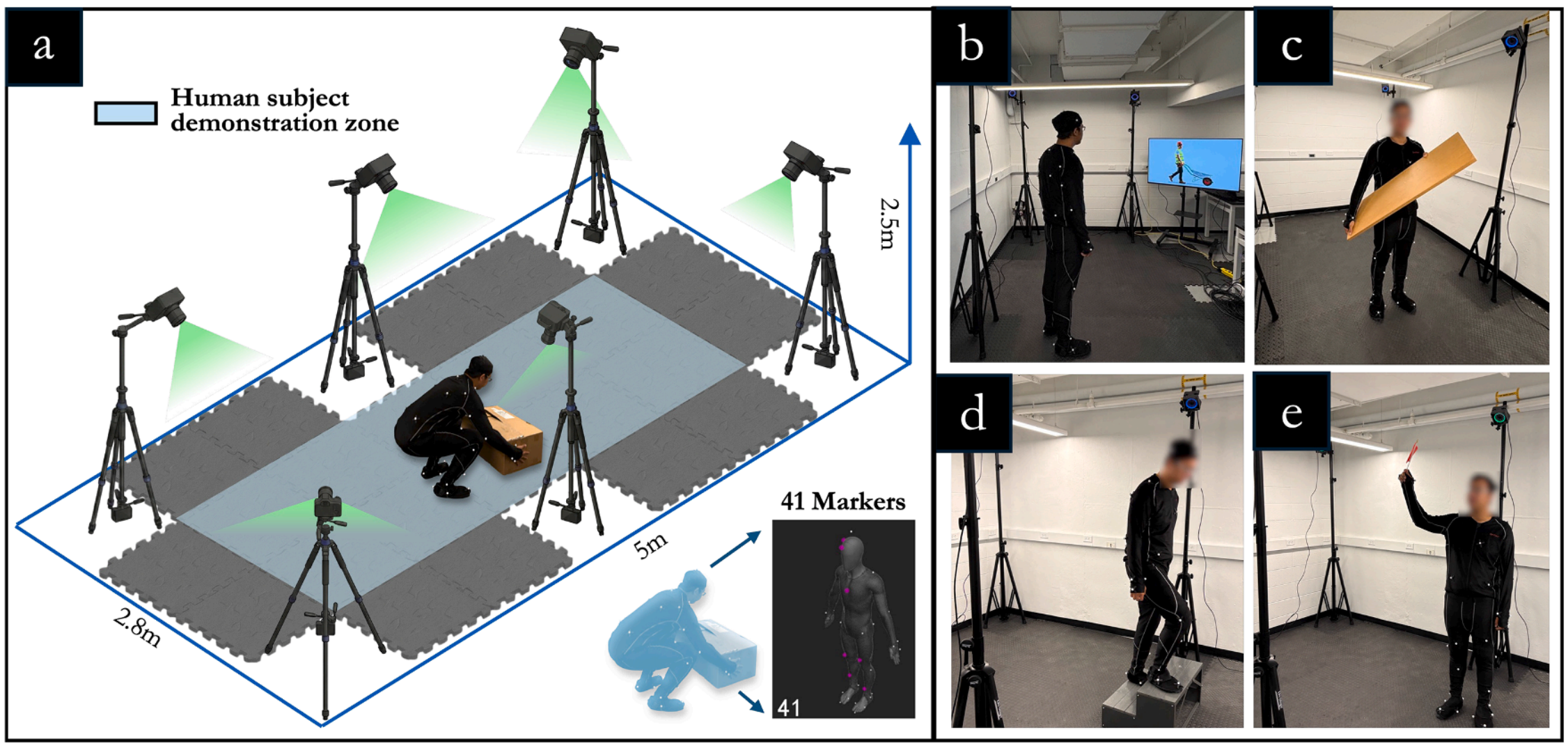

**Fig. 5.** Experimental setup for capturing human demonstrations of construction-related tasks: (a) layout of the data-collection workspace; (b) subject familiarization with each action prior to recording; (c) wood-carrying demonstration; (d) stair-climbing demonstration; and (e) flag-signaling demonstration.

**Table 2**
Construction-related actions performed by each subject for training the VPA system.

| No. | Action Name | No. | Action Name | No. | Action Name |
|---|---|---|---|---|---|
| 1 | Uneven-ground walking | 11 | Over-the-shoulder carrying | 21 | Material stacking |
| 2 | Foot probing | 12 | Back carrying | 22 | Wheelbarrow dragging |
| 3 | Jogging | 13 | Steel rod carrying (single hand) | 23 | Wheelbarrow pushing |
| 4 | Worksite pacing | 14 | Pipe carrying | 24 | Material dumping |
| 5 | Forward bending | 15 | Box lifting | 25 | Cable pulling |
| 6 | Turning with load | 16 | Wood panel carrying | 26 | Pipe rolling |
| 7 | Flagger signaling | 17 | Stair ascending | 27 | Heavy object tipping |
| 8 | Squat lifting | 18 | Stair descending | 28 | Painting |
| 9 | Overhead lifting & holding | 19 | Ramp ascending | 29 | Mortar cleaning-up |
| 10 | Concrete block carrying (two hands) | 20 | Set-in-place positioning | 30 | Rebar holding |

Notably, in this study, the 3D human pose estimation module of Humanoid-PoseNet was first pre-trained on the public Human3.6M dataset (Ionescu et al., 2014) (30,000 samples) and then fine-tuned on self-collected RGB-3D samples (9000 samples). Human3.6M is a widely accepted dataset for 3D human pose estimation. To ensure both the paired RGB-3D samples and the Human3.6M dataset have the same annotation style (the detected keypoints of the poses are the same), the authors mapped both the Human3.6M dataset (32 joints) and the self-collected RGB-3D samples (30 joints; Fig. 6-a) from OptiTrack skeleton to a 21-joint subset (Fig. 6-(b); hips, spine, thorax, neck, head, left shoulder, left arm, left forearm, left hand, right shoulder, right arm, right forearm, right hand, left up leg, left leg, left foot, left toe base, right up leg, right leg, right foot, right toe base). The resulting 21-joint 3D keypoints paired with RGB frames were used consistently for pre-training and fine-tuning. As 3D pose estimation is not the primary focus of this work, implementation details and training settings follow established protocols reported in the authors' prior work and representative 3D pose estimation studies (Liu & Jebelli, 2024; Pavllo et al., 2019).

For the human-to-humanoid pose retargeting module of Humanoid-PoseNet, the authors generated ground-truth humanoid pose targets from the captured 3D human poses (motion sequences) using a manual retargeting process adopted in humanoid motion imitation research (Cheng et al., 2024; Ji et al., 2025). Specifically, each captured 3D human pose was retargeted to the Unitree G1 humanoid skeleton by aligning source T-pose of the human and target T-pose of the humanoid robot, applying a predefined joint correspondence between the human skeleton and the humanoid kinematic chain, and compensating for coordinate-frame and body-scale differences between human and the humanoid robot. Here, the authors manually extracted and aligned the T-poses of the human and humanoid, and adjusted the body scale of the humanoid robot (set to 0.01 in this study) to ensure that body proportions could be correctly mapped to humanoid robots. The resulting retargeted motion was expressed in a compact 14-joint humanoid pose representation that matches the robot's kinematic structure. Fig. 6-(c) illustrates the retargeted 14 joints and the corresponding joint mapping between the human and humanoid configurations. Moreover, this dataset excludes non-actuated joints on the G1 platforms (e.g., the neck), ensuring that the translated poses are directly usable for control. Using the 9000 sampled frames, the authors obtained 9000 corresponding Unitree G1 humanoid pose targets which served as ground truth (i.e., $\mathscr{P}_r^{3D}$) for training the human-to-humanoid retargeting module described in Section 3.2 (with the paired 3D human poses as inputs). For both the paired RGB-3D samples and the retargeted humanoid pose samples, the authors used 80% of the data for model development (training and evaluation), with performance evaluated via 5-fold cross-validation. The remaining 20% was used as a test set to evaluate Humanoid-PoseNet. Results for 3D pose estimation and human-to-humanoid retargeting are reported in Section 5. Examples of the collected 3D human pose and the corresponding retargeted humanoid pose are shown in Fig. 6-(d) and -(e), respectively.

Finally, after training Humanoid-PoseNet, the authors applied it to generate retargeted humanoid pose trajectories for all 30 actions. Notably, to train the Humanoid-ActionNet, the input consisted of the retargeted motion sequences from each subject for each action trial (i.e., per-subject, per-action pose trajectories). That is, Humanoid-ActionNet was trained on time-ordered sequences of retargeted humanoid poses representing each demonstrated construction action. All retargeted poses were further mapped from the 14-joint representation to the 23-DoF G1 humanoid model to obtain the reference DoF positions used to construct humanoid states and compute reward terms. The reference DoFs include: left hip (pitch/roll/yaw), right hip (pitch/roll/yaw); left ankle (pitch/roll); right ankle (pitch/roll); left knee; right knee; left shoulder (pitch/roll/yaw); right shoulder (pitch/roll/yaw); left elbow; right elbow; waist (yaw/pitch/roll). All data was used to train the controller in IsaacGym using PPO within the teacher-student framework described in Section 3.3. Each RL episode sampled a reference motion segment and optimized the policy using the physics-aware reward functions to encourage accurate motion tracking, stable balance, and contact consistency. In addition, low-level control of the G1 humanoid robot was implemented using a joint-based PD controller (Equation.16), where the RL policy outputs target joint commands and the PD loop executes them in the IsaacGym simulation. The joint stiffness gains ($K_p$) of the PD controller were set to 200 N·m/rad for hip and knee joints, 300 N·m/rad for the torso joint, and 40 N·m/rad for ankle, shoulder, and elbow joints. The damping gains ($K_d$) were set to 5 N·m·s /rad for hip joints, 10 N·m·s/rad for knee joints, 6 N·m·s/rad for the torso joint, and 2 N·m·s/rad for ankle, shoulder, and elbow joints. The PPO policy learning rate was set to $10^{-4}$. Training was executed on a workstation equipped with one NVIDIA RTX PRO 6000 Blackwell Max-Q GPU. In this study, the policy converged after around 5000 iterations and was trained for 30,000 iterations to ensure stable convergence. All training results will be reported in Section 5.

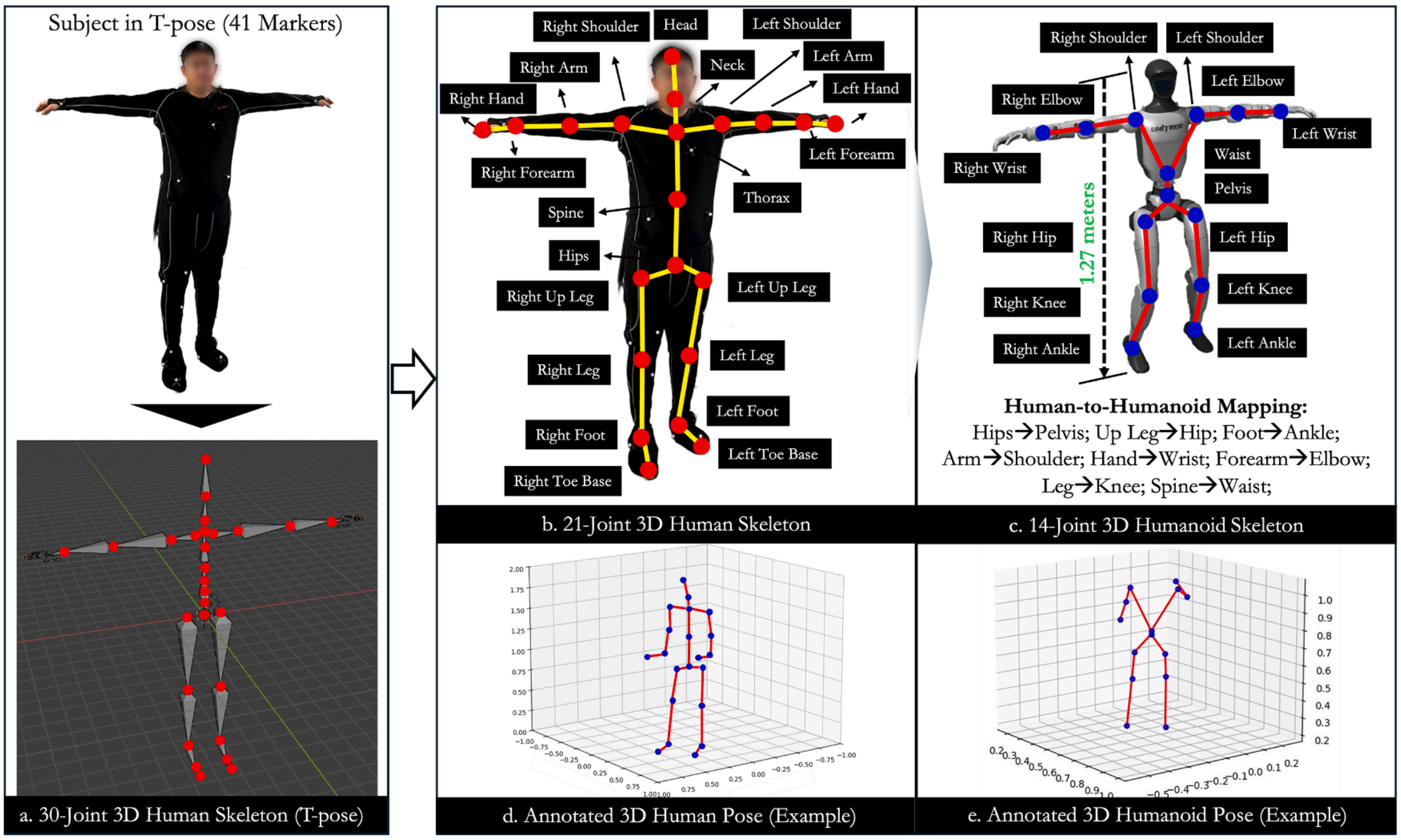

**Fig. 6.** Dataset annotation and pose representations used to train and evaluate Humanoid-PoseNet and to generate motion-sequence inputs for Humanoid-ActionNet: (a) OptiTrack 30-joint skeleton; (b) mapped 21-joint human pose; (c) 14-joint humanoid pose representation; (d) example annotated 3D human pose; and (e) example annotated 3D humanoid pose.

### *4.3. Testing session – retargeting accuracy and sim-to-sim/sim-to-real transfer on the Unitree G1 robot*

Once the Humanoid-PoseNet and Humanoid-ActionNet were trained, the integrated VPA system was evaluated for construction-related action execution through three stages. First, the human-to-humanoid retargeting performance was assessed on the held-out test set described in Section 4.2. Second, the VPA system was evaluated under a sim-to-sim process using two physics engines: IsaacGym (used for RL policy training) and MuJoCo (used as an independent simulator for evaluation). This cross-platform evaluation serves as a safety buffer, helping identify unstable motions caused by simulator-specific contact and dynamics modeling before hardware deployment (Erez et al., 2015). It also tests if the trained system transfers robustly across simulators under different dynamics implementations (Gu et al., 2024). Eight actions were selected from the full action set (Section 4.2) for sim-to-sim evaluation: (1) flagger signaling, (2) pushing a wheelbarrow, (3) over-the-shoulder carrying, (4) carrying steel rod, (5) carrying concrete block, (6) dragging a wheelbarrow, (7) carrying wood, and (8) carrying pipe. In this study, actions that did not pass the sim-to-sim evaluation were not deployed on the physical robot (Unitree G1) to reduce hardware risk and protect the on-site human supervisor. Only actions that passed the sim-to-sim evaluation proceeded to the sim-to-real testing. Third, for each action that passed sim-to-sim testing, the sim-to-real transfer was tested on a physical Unitree G1 humanoid robot (Fig. 7). As shown, the physical G1 matched the simulated robot in actuation layout, joint limits, and kinematic structure. Minor differences existed in the hand configuration between hardware and simulation; however, this study focuses on whole-body humanoid locomotion rather than dexterous manipulation. As a result, the hand-level mismatch did not affect policy training or the evaluation used in this work. Hardware deployment followed Unitree's Physical Deployment Guide (Unitree Robotics, 2024). A calibration step was performed to align joint zero positions and kinematic parameters between simulation and hardware. The G1 was then controlled by the VPA system to reproduce motion sequences for the selected tested actions mentioned above.

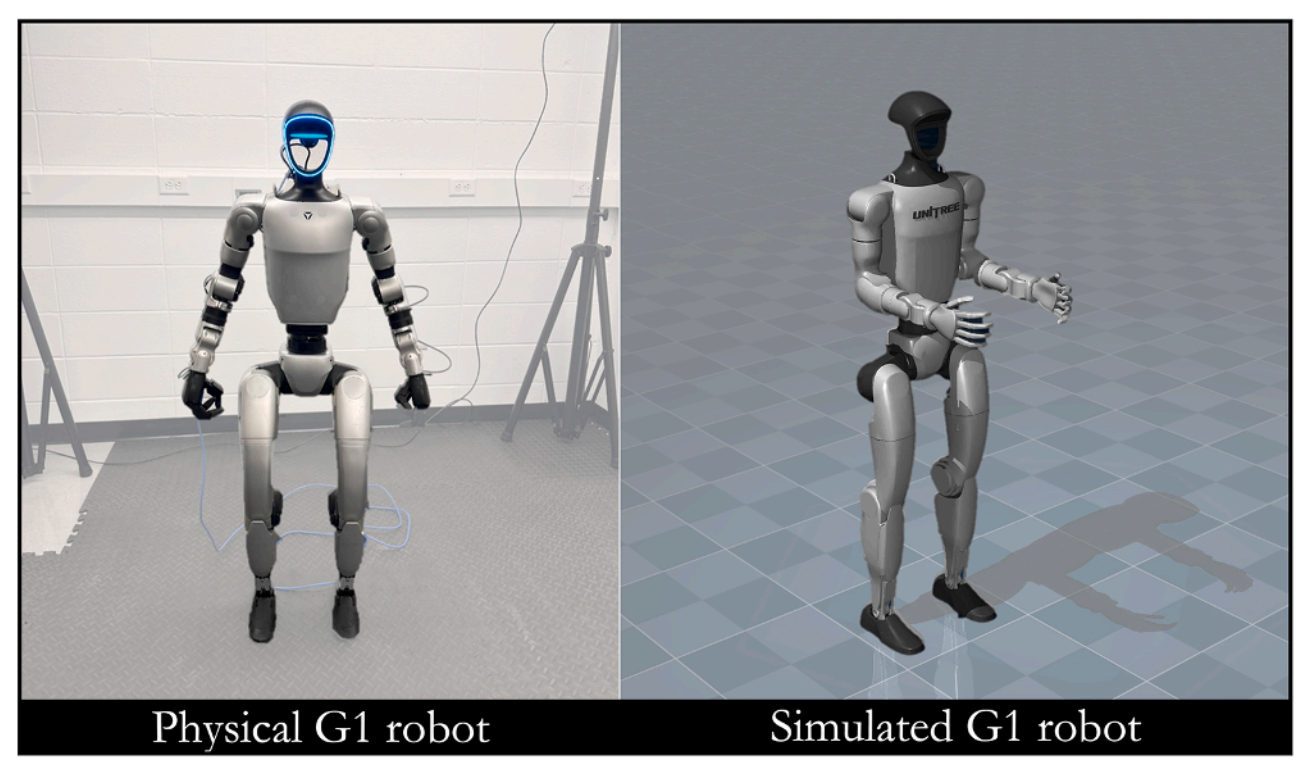

**Fig. 7.** Simulated and physical Unitree G1 humanoid robot used for sim-to-sim and sim-to-real evaluation.

For evaluation, retargeting accuracy was measured using Mean Per Joint Position Error (MPJPE) relative to the ground-truth retargeted humanoid poses. For both sim-to-sim and sim-to-real transfer, MPJPE was likewise used to quantify the deviation between the reference pose trajectories generated by Humanoid-PoseNet and the executed pose trajectories generated by the RL whole-body controller (i.e., $\pi_{\text{student}}$), consistent with established evaluation protocols in previous humanoid control studies (He, Luo, He et al., 2024; Ji et al., 2025). Testing results and examples for all three evaluations are reported in Section 5.

## 5. Results

This section reports evaluation results for the proposed VPA system, including (i) the human-to-humanoid pose retargeting performance of Humanoid-PoseNet and (ii) the whole-body control performance of Humanoid-ActionNet for construction-related task execution in simulation and on a physical Unitree G1 humanoid robot.

### 5.1. *Performance of Humanoid-PoseNet in human-to-humanoid pose translation*

Humanoid-PoseNet was first trained to extract human 3D poses from RGB inputs. To enable this capability, the authors used an off-the-shelf 3D pose estimation model, PoseNet (Moon et al., 2019), as the first part of the Humanoid-PoseNet. The model was fine-tuned using the dataset described in Section 4.2, and its performance – quantified by MPJPE on the test set – together with its outputs, is reported in Fig. 8. Here, PoseNet was selected based on a comparative evaluation against AlphaPose, VideoPose3D, and the authors' prior model (3D-PoseNet) (Fang et al., 2022; Liu & Jebelli, 2024; Pavllo et al., 2019). Notably, the full model comparative results are reported and discussed in the Discussion section (Section 6), with detailed results provided in Table 5. Briefly, PoseNet achieved the lowest MPJPE among the evaluated models, with an average MPJPE of 58.3 mm, outperforming the second-best model by 2 mm. Next, the human-to-humanoid retargeting module was evaluated. The network was trained using the hyperparameter settings introduced in Section 3.2, and the total training loss $\mathscr{L}_{\text{total}}$ (Equation.5) was used to quantify convergence during training. For evaluation, each estimated human pose $\mathscr{P}_h^{3D}$ from the test set (Section 4.3) was used as input, and the network generated the corresponding retargeted humanoid pose $\widehat{\mathscr{P}}_r^{3D}$. This retargeted pose was then compared with the paired ground-truth humanoid pose $\mathscr{P}_r^{3D}$ to compute the pose-level deviation in the humanoid configuration space, measured by the MPJPE of humanoid 3D keypoints. Table 3 summarizes the evaluation performance of the human-to-humanoid retargeting module. Overall, for the test data, the proposed method achieved an average MPJPE of 48.46 mm for all 14 joints. Fig. 8 further visualizes representative retargeting examples for four construction-related actions, showing that the proposed mapping preserves key posture patterns while maintaining humanoid-compatible joint configurations. For each example, the corresponding error (i.e., the MPJPE error relative to the ground truth) is also reported. In addition, Fig. 9 presents a time-series example of the carrying-pipe action, illustrating retargeting performance across a full motion sequence rather than isolated frames. As shown, the retargeted humanoid poses closely mimic the inter-joint

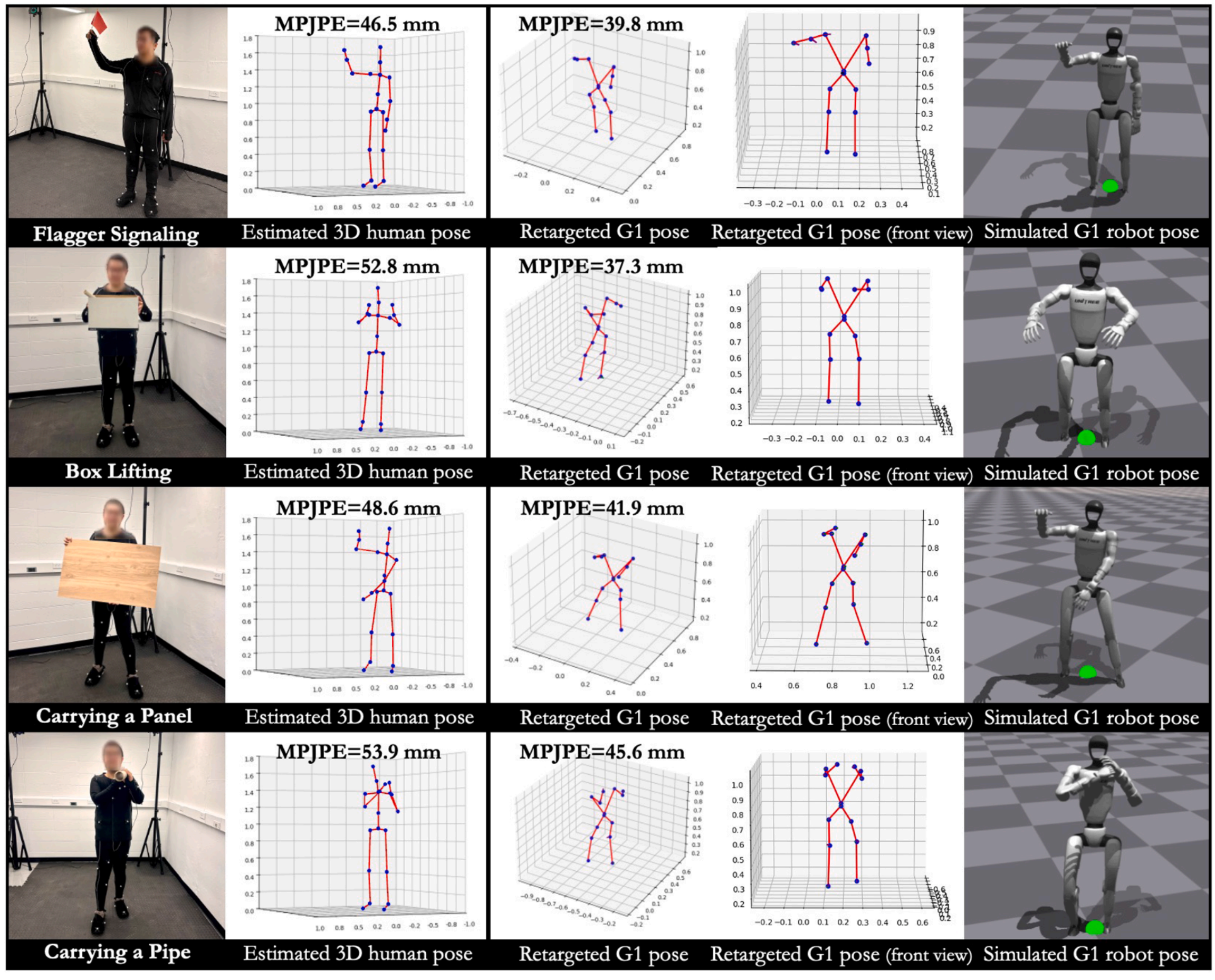

**Fig. 8.** Performance of Humanoid-PoseNet in 3D human pose estimation and 3D human-to-humanoid retargeting.

**Table 3**
Performance of the Humanoid-PoseNet for human-to-humanoid retargeting on the testing dataset.

| Joints<br>Model | Arm | Forearm | Hand | Spine | Hip | Up Leg | Leg | Foot | MPJPE*<br>(14 joints) |
|---|---|---|---|---|---|---|---|---|---|
| **PoseNet** + HtH retargeting | 41.4 | 49.2 | 47.9 | 50.5 | 45.9 | 50.1 | 51.8 | 50.6 | 48.46 |

*Note: MPJPE reported for Arm, Forearm, Hand, Upper Leg, Leg, and Foot are the average of the left and right sides.

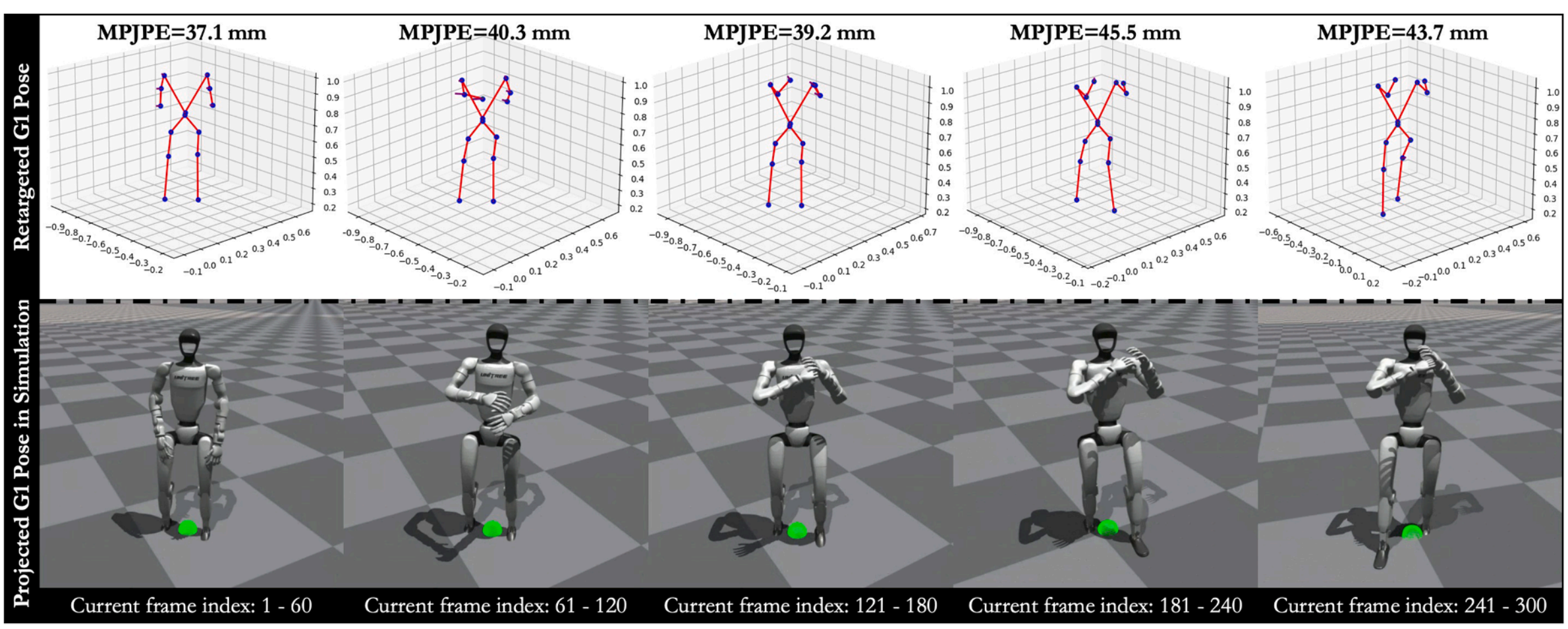

**Fig. 9.** Example human-to-humanoid retargeting result for a construction-related action (carrying-pipe).

coordination of human postures while remaining consistent with the Unitree G1's kinematic configuration. Overall, these results indicate that Humanoid-PoseNet can translate diverse worker postures into feasible humanoid configurations, providing reliable reference trajectories for the following action learning stage. Moreover, the ablation study of the human-to-humanoid retargeting module will be reported in the Discussion section.

### 5.2. *Performance of Humanoid-ActionNet in construction actions execution*

Next, the authors report the training and evaluation performance of Humanoid-ActionNet. As described earlier, Humanoid-ActionNet adopts a teacher-student RL framework in which the teacher policy, $\pi_{\text{teacher}}$, was trained with PPO and the student policy, $\pi_{\text{student}}$, was trained via distillation. The teacher policy $\pi_{\text{teacher}}$ was trained with privileged observations to maximize task rewards while enforcing physical feasibility (balance, contact consistency, and motion tracking). The student policy $\pi_{\text{student}}$ was then distilled to operate using only deployable sensing, replacing privileged signals with a short history of proprioceptive states (window size $w = 10$). Training curves in Fig. 10 show stable convergence of the PPO objectives for the teacher policy, including the surrogate loss $\mathscr{L}_{\text{actor}}$ (Equation.11) and the value loss $\mathscr{L}_{\text{critic}}$ (Equation.13). After training 30,000 iterations, the surrogate loss stabilized around $-0.0037$ (unitless) and the value loss decreased to 0.061 (unitless). In addition, $\pi_{\text{student}}$ closely matched $\pi_{\text{teacher}}$ through distillation loss of 0.0056 (unitless).

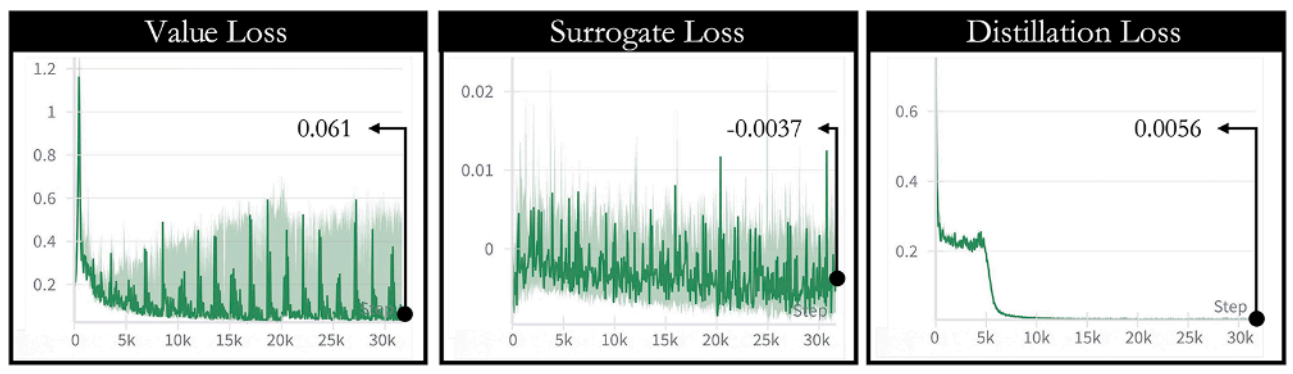

**Fig. 10.** Training performance of the teacher policy and distilled student policy in Humanoid-ActionNet.

During evaluation, Humanoid-ActionNet deployed the student policy, $\pi_{\text{student}}$, to control the Unitree G1 humanoid robot and reproduce the pose trajectories generated by Humanoid-PoseNet while maintaining stable bipedal locomotion and coherent upper-body manipulation patterns under contact interactions. Evaluation was first conducted in a simulation. Specifically, $\pi_{\text{student}}$ was tested in IsaacGym under the same motion-tracking and task settings used during training, and was then evaluated for sim-to-sim transfer in MuJoCo. Eight tested construction-related actions (refer to Section 4.3) were used for evaluation. Fig. 11 illustrates an example of each action execution and the cross-simulator transfer from IsaacGym to MuJoCo. As shown, using $\pi_{\text{student}}$, the G1 successfully executed all eight actions in simulation and maintained consistent performance across the two simulation environments. In addition, Table 4 reports the per-joint MPJPE for each action; across all eight actions, the average MPJPE was 82.45 millimeters (mm).

Finally, the authors deployed the trained Humanoid-PoseNet and Humanoid-ActionNet on a physical Unitree G1 humanoid robot to evaluate sim-to-real transfer. The G1 successfully reproduced all eight tested actions in the physical environment, and its real-world performance was consistent with the sim-to-sim evaluation trends observed in IsaacGym-to-MuJoCo transfer. Fig. 12 presents several representative deployment results, showing the G1 performing three construction-related actions (flagger signaling, carrying a pipe, and carrying a concrete block) while maintaining stable balance throughout execution.

## 6. Discussion

According to the reported results, the proposed VPA system demonstrates the potential to enable humanoid robots to (1) retarget vision-based worker demonstrations into humanoid-compatible, pose-driven motions and (2) execute these motions under dynamics and contact constraints in simulation and physical environments. Below, the authors further discuss several key technical insights of the proposed system.

First, as described in both Section 3.2 and Section 5.1, the 3D pose estimation module in Humanoid-PoseNet for extracting human posture from RGB inputs can be configured using off-the-shelf methods evaluated in previous 3D human pose estimation studies. To examine whether

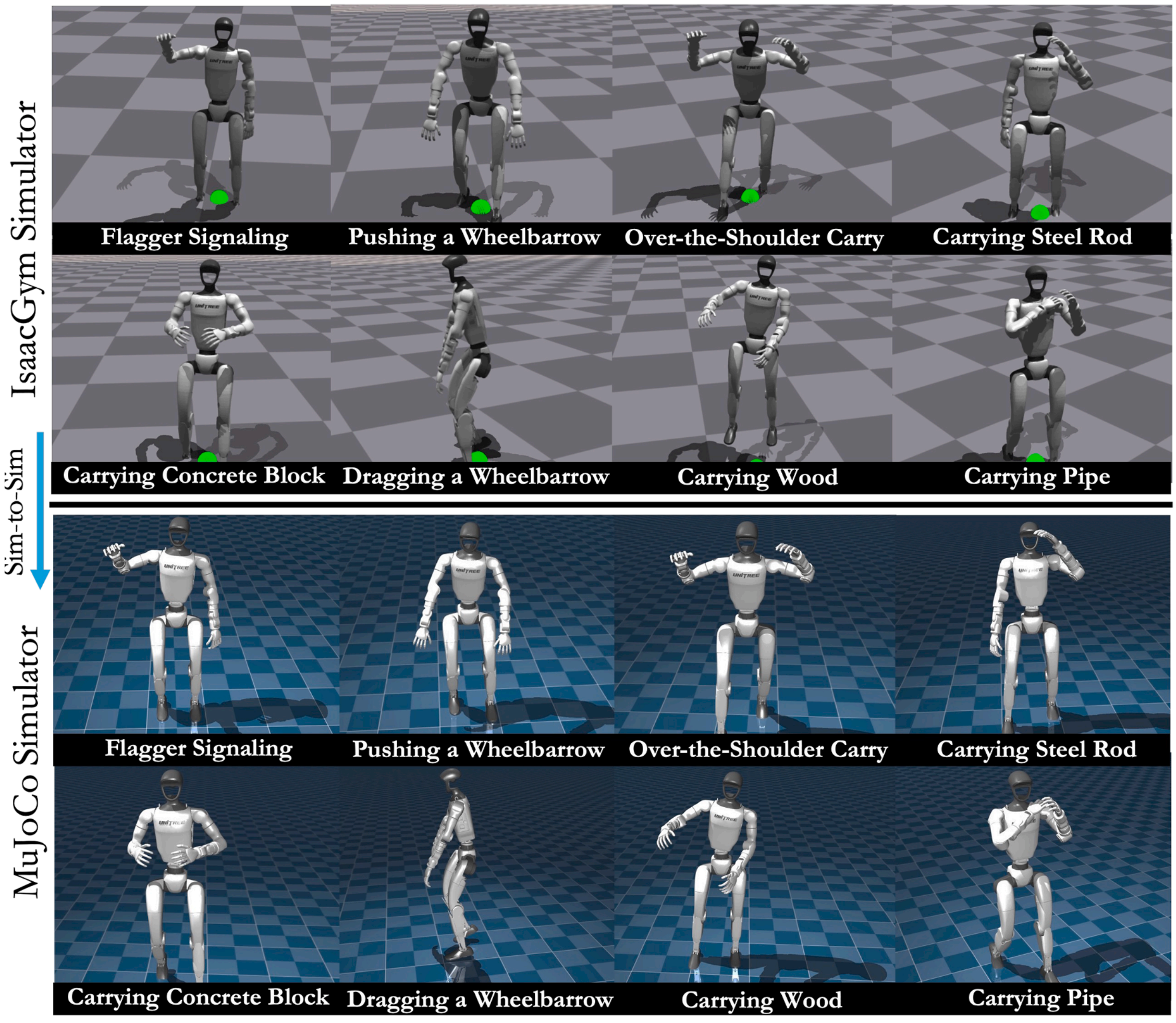

**Fig. 11.** Sim (IsaacGym)-to-sim (MuJoCo) evaluation of the distilled student policy on the Unitree G1 across eight tested construction-related actions.

**Table 4**
Per-joint MPJPE of the Humanoid-ActionNet across all eight tested actions.

| Joints<br>Action | Arm | Forearm | Hand | Spine | Hip | Up Leg | Leg | Foot | MPJPE*<br>(14 joints) |
|---|---|---|---|---|---|---|---|---|---|
| Flagger signaling | 109.2 | 98.2 | 86.2 | 65.7 | 50.9 | 80.6 | 83.7 | 78.2 | 84.91 |
| Pushing a wheelbarrow | 88.7 | 89.3 | 57.7 | 66.9 | 60.7 | 96.2 | 88.6 | 90.7 | 82.14 |
| Over-the-shoulder carry | 121.4 | 85.8 | 67.9 | 70.5 | 50.9 | 90.1 | 92.8 | 90.6 | 87.04 |
| Carrying steel rod | 93.2 | 91.6 | 61.7 | 62.1 | 55.2 | 83.4 | 85.6 | 83.4 | 79.65 |
| Carry concrete block | 82.3 | 88.1 | 65.2 | 69.6 | 50.9 | 87.1 | 93.7 | 88.2 | 80.69 |
| Dragging a wheelbarrow | 88.7 | 89.7 | 58.1 | 71.2 | 60.7 | 95.1 | 91.3 | 88.4 | 82.46 |
| Carrying wood | 93.3 | 85.9 | 70.3 | 59.2 | 49.8 | 87.8 | 93.4 | 93.9 | 82.73 |
| Carrying pipe | 90.6 | 85.8 | 60.3 | 63.9 | 53.2 | 85.4 | 89.6 | 89.4 | 79.95 |
| **Average** | 95.93 | 89.30 | 65.93 | 66.14 | 54.04 | 88.21 | 89.84 | 87.85 | **82.45** |

*Note: MPJPE reported for Arm, Forearm, Hand, Upper Leg, Leg, and Foot are the average of the left and right sides.

network selection would affect the performance of the entire pipeline, the authors compared multiple available representative models using the same dataset, annotation protocol, and evaluation settings described in Section 4. All these models (i.e., PoseNet, AlphaPose, VideoPose3D, and 3D-PoseNet) are widely applied to pose estimation applications, as reported in (Fang et al., 2022; Liu & Jebelli, 2024; Moon et al., 2019; Pavllo et al., 2019). In other words, all four models were fine-tuned and tested using the same human pose training and test datasets reported in Section 4.1. The resulting 3D poses were then passed into the same human-to-humanoid retargeting module in Humanoid-PoseNet with the same architecture and hyperparameters across all comparisons. Table 5 summarizes human pose estimation performance on the Section 4 test set using MPJPE. Table 6 reports the corresponding retargeting results using the same metric (i.e., MPJPE for humanoid pose) as in Table 3. For both tables, lower values indicate better performance. As shown in Table 5, PoseNet achieved the best pose-estimation performance among the evaluated models, with the lowest MPJPE. This result supports the selection of PoseNet as the 3D pose-estimation backbone in the Results section (Section 5.1). In addition, Table 6 shows that the downstream retargeting results were generally comparable across different pose-estimation backbones, suggesting that the proposed VPA system is not tightly coupled to a specific 3D pose-estimation model.

Second, the authors conducted an ablation study to better understand the contribution of each loss function for the human-to-humanoid retargeting module (Section 3.2). To be more specific, one loss function (i.e., $\mathscr{L}_1$, $\mathscr{L}_2$, or $\mathscr{L}_3$) was removed from the retargeting module at a

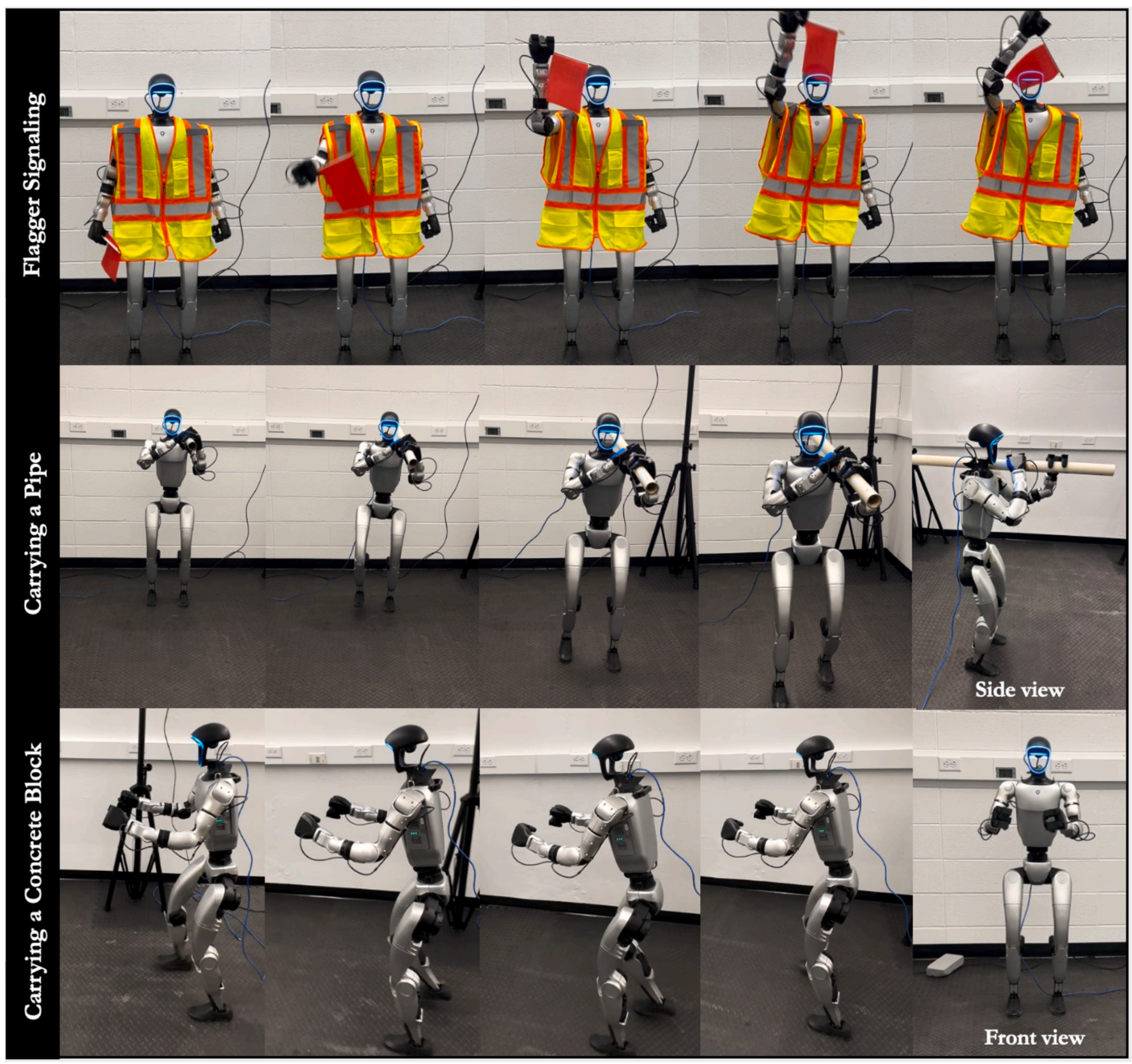

**Fig. 12.** Real-world deployment of the proposed VPA system on the Unitree G1 robot for execution of three construction-related actions*.
*Note: For the flagger-signaling trial, the robot wore a high-visibility reflective vest following common site practice. For the concrete-block trial, because the deployed G1 platform is equipped with a three-finger hand that is not suited for securely grasping a block, the results are reported only as whole-body posture execution.

**Table 5**
Human pose estimation performance comparison across competing 3D pose-estimation models.

| Model | MPJPE (all 21 joints) |
|---|---|
| AlphaPose | 60.6 |
| VideoPose3D | 60.3 |
| **PoseNet** | **58.3** |
| 3D-PoseNet | 61.0 |

time, while keeping the same training procedure, dataset, and parameter settings to ensure a fair comparison. For each ablation setting, retargeting quality was evaluated using MPJPE relative to the ground-truth humanoid pose. The results (Table 7) show that removing each loss component degraded retargeting precision by increasing MPJPE by 57.91 mm, 37.33 mm, and 8.76 mm, respectively, compared with the retargeting module that uses all three loss components. These results indicate that each loss contributes to the overall retargeting performance of the retargeting network. From a mechanism perspective, $\mathscr{L}_1$ is designed to enforce a geometry-aware alignment of the shared latent

**Table 6**
Per-joint analysis for human-to-humanoid retargeting when using different 3D pose-estimation models with the same retargeting module as introduced in Section 3.2.

| Joints<br>Model | Arm | Forearm | Hand | Spine | Hip | Up Leg | Leg | Foot | MPJPE*<br>(14 joints) |
|---|---|---|---|---|---|---|---|---|---|
| AlphaPose + HtH retargeting** | 43.2 | 48.1 | 56.2 | 40.7 | 45.9 | 56.6 | 53.7 | 48.2 | 49.90 |
| VideoPose3D + HtH retargeting | 49.7 | 50.3 | 51.7 | 46.9 | 47.7 | 52.2 | 48.6 | 50.7 | 50.07 |
| **PoseNet** + HtH retargeting | 41.4 | 49.2 | 47.9 | 50.5 | 45.9 | 50.1 | 51.8 | 50.6 | **48.46** |
| 3D-PoseNet + HtH retargeting | 51.2 | 50.6 | 55.3 | 50.1 | 45.2 | 51.4 | 49.6 | 45.4 | 50.16 |

*Note: MPJPE reported for Arm, Forearm, Hand, Upper Leg, Leg, and Foot are the average of the left and right sides.
**Note: HtH retargeting stands for human-to-humanoid retargeting module in Humanoid-PoseNet.

**Table 7**
Ablation study of the human-to-humanoid retargeting module in Humanoid-PoseNet.

| $\mathscr{L}_1$ loss in Equation.5 | $\mathscr{L}_2$ loss in Equation.5 | $\mathscr{L}_3$ loss in Equation.5 | MPJPE (14 joints) |
|---|---|---|---|
| excluded | included | included | 106.37 |
| included | excluded | included | 85.79 |
| included | included | excluded | 57.22 |
| included | included | included | 48.46 |

space by grouping pose pairs with similar BAE across the human and humanoid domains. Without $\mathscr{L}_1$, the network loses the cross-skeleton similarity index, greatly reducing its ability to link similar human and humanoid postures and retargeting precision. Upon $\mathscr{L}_1$, the latent-consistency loss $\mathscr{L}_3$ further enhances cross-domain distribution alignment by encouraging embeddings generated from human poses to remain consistent after decoding and re-encoding in the humanoid domain. Therefore, removing $\mathscr{L}_3$ still reduces retargeting precision, but with a smaller impact than removing $\mathscr{L}_1$, as it acts as a stabilizing constraint rather than the main alignment driver. $\mathscr{L}_2$ acts as an anchor that ties the humanoid robot-side encoder and decoder to valid humanoid pose configurations by enforcing accurate reconstruction on robot-domain inputs (i.e., ground-truth humanoid pose); removing $\mathscr{L}_2$ increases the likelihood of robot-infeasible outputs, thereby reducing the retargeting precision.

In addition to this ablation study, the adaptability of the human-to-humanoid retargeting module to alternative humanoid configurations was also examined. Using the same pipeline, the retargeting model was trained and evaluated using the Unitree H1 kinematic configuration as the target domain rather than the Unitree G1. Fig. 13 presents representative retargeting results for the H1 (carrying-wood action) and the G1 (dragging-a-wheelbarrow action). In each example, the retargeted joint keypoints – shown as red and green dots – are obtained from the corresponding human poses and projected onto the robot's joint locations in the target configuration space. The results show that the module can reliably retarget human pose trajectories not only to the G1 but also to the H1, indicating promising adaptability of the proposed retargeting approach.

Third, beyond Humanoid-PoseNet, the performance of Humanoid-ActionNet reported in Section 5.2 shows that the humanoid robot can maintain balance and execute construction-related behaviors using the RL policy. In addition to the reliable PPO-based training performance reported in Fig. 10, a major factor contributing to this reliable execution is the physics-aware reward design. The reward terms were constructed to jointly encourage (1) accurate motion tracking based on the retargeted humanoid pose, (2) dynamic feasibility (e.g., balance, smoothness, and contact integrity), and (3) task-oriented action consistency for construction behaviors. These terms guide policy optimization toward stable control strategies that enable the humanoid robot to maintain its whole-body balance to perform the tasks.

To further evaluate Humanoid-ActionNet, this study conducted a comparative analysis with representative whole-body humanoid control methods, including OmniH2O (He, Luo, He et al., 2024), ExBody (Cheng et al., 2024), ExBody with Adversarial Motion Priors (AMP) (Cheng et al., 2024; Peng et al., 2021), and ExBody2 (Ji et al., 2025). Here, all baseline methods were fine-tuned using the same retargeted humanoid motion dataset used for Humanoid-ActionNet and were evaluated on the same eight construction-related actions reported in Table 4. The same MPJPE metric was used to quantify the deviation between the reference and executed humanoid pose trajectories. As reported in Table 8, Humanoid-ActionNet achieved the lowest average MPJPE among the evaluated methods. This result indicates that the proposed controller tracked the construction-related retargeted motions more accurately than the compared baselines under the evaluation setting of this study. One possible explanation is that Humanoid-ActionNet was designed around the task characteristics of construction-related motions. Construction-related actions often involve task-specific postures, load-bearing movement patterns, and coordinated upper- and lower-body motions. The reward terms in Humanoid-ActionNet, as reported in Table 1, were designed to better support these requirements. In contrast, existing whole-body control methods such as ExBody, ExBody2, and OmniH2O were mainly developed for general locomotion and expressive whole-body motion (e.g., dancing). Therefore, while these methods provide strong baselines for humanoid whole-body control, their original design objectives are not fully aligned with construction-related task execution. Overall, this comparison further supports the practical value of Humanoid-ActionNet within the proposed VPA system. By combining retargeted construction-related motion references with physics-aware reward design and teacher-student policy learning, Humanoid-ActionNet enables the humanoid robot to more reliably convert worker demonstrations into physically executable whole-body actions.

Despite the technical merits of the RL-based whole-body robotic controller and the promising task execution performances observed in sim-to-sim and sim-to-real transfers visualized in Figures.11 and 12, the authors acknowledged that the trained robot control policy deployment transfers also experience several failure cases during sim-to-sim transfer process. Notably, to mitigate hardware risk, the trained RL policies, $\pi_{\text{student}}$, were required to pass sim (IsaacGym)-to-sim (MuJoCo) evaluation before being deployed to the physical Unitree G1. Fig. 14 illustrates several representative failure cases in MuJoCo, including loss of balance (Fig. 14-(a)), unstable recovery after perturbations (Fig. 14-(b)), and tracking breakdown (i.e., the humanoid robot could not mimic the human demonstration; Fig. 14-(c)). These failures can be attributed to: (a) the difficulty of fully capturing contact-rich construction dynamics through the current reward functions, and (b) dynamics and actuation mismatches across simulation engines (e.g., IsaacGym vs. MuJoCo), as well as the additional gap between simulation and real hardware actuation (Acosta et al., 2022; Erez et al., 2015). How to systematically improve robustness across simulators and strengthen contact-sensitive reward shaping (particularly for construction tasks) remains an open-ended question, and should be explored in future work.

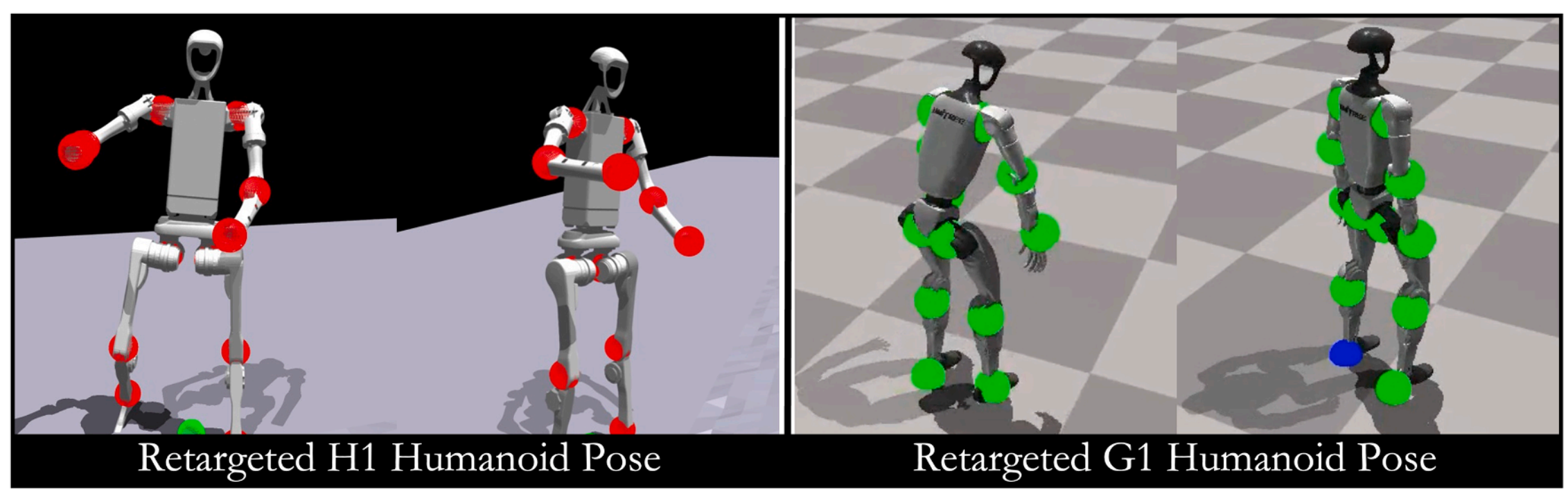

**Fig. 13.** Human-to-humanoid pose retargeting results for Unitree H1 and G1 humanoid robots.

**Table 8**
Comparative evaluation of Humanoid-ActionNet and representative whole-body humanoid control baselines.

| Joints<br>Model | Arm | Forearm | Hand | Spine | Hip | Up Leg | Leg | Foot | MPJPE (14 joints) |
|---|---|---|---|---|---|---|---|---|---|
| ExBody | 123.4 | 113.2 | 101.8 | 79.1 | 69.5 | 105.6 | 100.7 | 100.3 | 102.75 |
| ExBody + AMP | 98.3 | 93.7 | 74.5 | 67.8 | 61.6 | 97.2 | 91.8 | 111.1 | 90.19 |
| OmniH2O | 116.6 | 92.4 | 78.2 | 70.4 | 59.6 | 99.7 | 96.3 | 126.3 | 96.35 |
| ExBody2 | 101.8 | 93.6 | 69.4 | 61.9 | 55.1 | 95.9 | 84.5 | 108.8 | 87.48 |
| **Humanoid-ActionNet** | 95.9 | 89.3 | 65.9 | 66.1 | 54.0 | 88.2 | 89.8 | 87.9 | **82.45** |

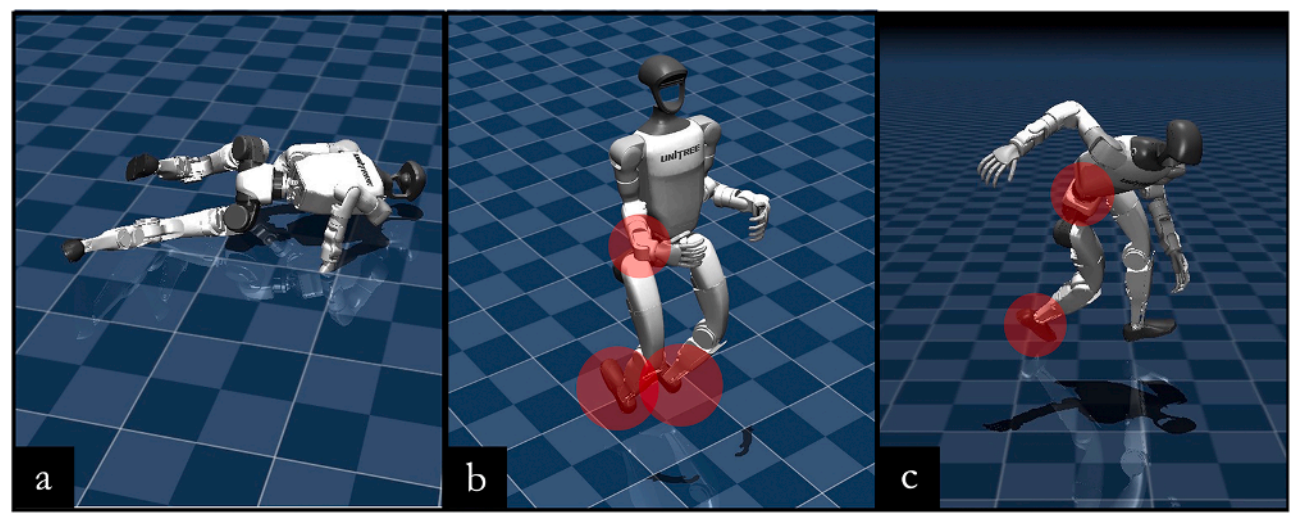

**Fig. 14.** Representative failure cases observed in MuJoCo during sim-to-sim evaluation.

Based on the quantitative evaluation of the proposed VPA pipeline, the humanoid robot showed promising capability in executing selected construction-related tasks. These results suggest that humanoid robots may offer a complementary direction within construction robotics, particularly for worker-like tasks that require human-compatible movement patterns, whole-body coordination, multi-task adaptability, and operation in human-designed spaces. However, construction tasks vary widely in their requirements, and different robotic platforms may be better suited to different applications (Chen et al., 2022, 2025; Asadi et al., 2021; Rakha et al., 2018; Sheng et al., 2025). Table 9 summarizes the representative advantages, deployment considerations, and typical construction applications of humanoid robots compared with other robotic platforms used in construction, including Unmanned Aerial Vehicles (UAVs), Unmanned Ground Vehicles (UGVs), robotic arms, and quadruped robots. This comparison is not intended to provide an absolute ranking of robotic platforms. Instead, it highlights their different potential roles in construction robotics. UAVs are valuable for aerial perception and inspection, UGVs are useful for ground-level transport and monitoring, robotic arms are effective for structured and repetitive manipulation, and quadruped robots are suitable for rough-terrain inspection and sensing. In comparison, humanoid robots may provide additional value for worker-like construction tasks that require multi-task adaptability and compatibility with human-designed spaces, tools, and workflows.

**Table 9**
Comparison of representative robotic platforms for construction applications.

| Robot Platform | Representative Advantages | Deployment Considerations | Typical Construction Applications |
|---|---|---|---|
| UAVs (Rakha & Gorodetsky, 2018; Zhao et al., 2026) | Rapid aerial coverage of large or elevated areas | Limited manipulation capability | Aerial inspection, mapping, progress monitoring |
| UGVs (Chen et al., 2022) | Stable ground mobility | Limited by stairs, clutter, and narrow paths | Ground inspection, monitoring, material delivery |
| Robotic Arms (Asadi et al., 2021) | Precise, repeatable manipulation | Requires structured workspace and fixed or mobile base setup | Prefabrication, welding, bricklaying, assembly |
| Quadruped Robots (Chen et al., 2025b) | Strong uneven-terrain mobility | Limited manipulation capability | Inspection, mapping, monitoring in uneven terrain |
| Humanoid Robots (Sheng et al., 2025) | Human-like whole-body coordination and manipulation | Balance and whole-body control challenges (explored in this study) | Has potential to perform worker-like task execution and multi-task transitions |

Overall, according to the above discussion, the results of this study indicate that the integrated pipeline has the potential to enable humanoid robots to execute diverse construction-related tasks in a worker-like manner, involving varied postures and motions. This task adaptability differentiates humanoid robots from many task-specific robotic systems currently deployed or studied in construction, which are often designed for a limited set of predefined operations (e.g., pick-and-place operations) (Duan et al., 2025; Radosavovic et al., 2024). In addition, as discussed, the proposed system can be applied to humanoid robots with different configurations (e.g., Unitree H1 and G1), suggesting that the proposed pipeline is not limited to a single platform and can potentially be adapted to other humanoid systems (e.g., Boston Dynamics' Atlas and Tesla's Optimus) that have the mechanical strength to perform construction-related tasks. Taken together, the proposed system offers a promising pathway for extending humanoid learning to a broader set of humanoid robots and construction tasks in human-centric workspaces, supporting scalable humanoid-assisted workflows and accelerating construction automation.

## 7. Conclusion

This study designed a Vision-based Perception-and-Action (VPA) system that innovatively equips a humanoid robot with two new functions: learning humanoid poses from construction workers and converting these poses into robot-executable actions for construction task execution. Relying on RGB data from worker demonstrations, the VPA system reconstructs 3D human motion and retargets it into humanoid-compatible pose trajectories, addressing the human-humanoid morphology mismatch during the retargeting process. Given the retargeted poses, the system further leverages a physics-aware reinforcement learning structure to generate stable actions that can be executed by the robot under contact and balance constraints, bridging a key gap in developing humanoid controllers for construction-related task execution. A series of construction-related motions was performed by human subjects to train the robot and evaluate the feasibility of the proposed VPA system. Results showed that a humanoid platform (Unitree G1) can reproduce these construction behaviors with a reliable task success rate while maintaining stability in the physical world. In sum, this study contributes new knowledge to construction robotics by developing a learning-to-execution pipeline for humanoid robots to execute construction tasks. Moreover, the proposed system offers a foundational step toward practical humanoid deployment in civil and construction engineering, opening new opportunities for next-generation automation in human-dominated jobsites.

This work also has several limitations that direct future studies. First, the robustness of the proposed VPA system is constrained by the limited number of human subjects and the scope of activities included in the dataset collected in this study. To better evaluate and enhance the generalizability of the proposed system, future studies can incorporate existing construction activity datasets (e.g., CML: construction motion data library (Tian et al., 2022)) or collect additional datasets to prepare larger construction worker activity datasets involving workers with different experience levels, body characteristics, construction tasks, and task execution styles. Such datasets would help examine how variations in worker motion patterns influence humanoid pose retargeting and action learning, and would further improve the robustness of the proposed VPA system across diverse worker populations and construction scenarios. Second, the current system focuses on posture-based motion imitation and does not model humanoid interaction with construction tools or materials, which is required for a wide range of construction tasks. Future studies should investigate how to integrate tool and material interaction into the humanoid's learning-to-execution pipeline (i.e., VPA system). Third, although the designed reward functions enable reliable task execution, the sim-to-sim transfer still showed the non-negligible failure cases across the eight actions. Future work should refine the reward design, such as improving contact consistency, balance margins, and action smoothness, to enhance execution robustness and reduce sim-to-sim and sim-to-real gaps. Future studies can also explore realistic simulation techniques, such as 3D Gaussian Splatting (3DGS) (Cheng et al., 2025; Zhou et al., 2024), to improve the visual realism of construction environments used for humanoid robot training. For example, 3DGS-based reconstruction could be used to create more realistic jobsite scenes and better represent spatial context during humanoid policy learning and testing, thereby helping reduce the visual and geometric components of the sim-to-real gap. To further address the physical interaction demands of construction tasks, 3DGS-based scene reconstruction should be combined with physics-calibrated modeling of contact, friction, actuation, payload, and material properties to improve overall sim-to-real transfer for humanoid construction task execution. Finally, the experiments in this study focus on short-horizon behaviors (i.e., single actions). This setting cannot address the requirements of long-horizon, multi-step construction tasks (e.g., building a wall), making the extension to sequential task execution an important direction for future humanoid robot research.

## CRediT authorship contribution statement

**Yanxi Liu:** Writing – original draft, Visualization, Validation, Methodology, Formal analysis, Data curation, Writing – review & editing. **Yizhi Liu:** Writing – original draft, Visualization, Supervision, Methodology, Funding acquisition, Conceptualization, Writing – review & editing.

## Declaration of competing interest

The authors declare that they have no known competing financial interests or personal relationships that could have appeared to influence the work reported in this paper.

## Acknowledgement

The work was supported by an NVIDIA Academic Grant. Any opinions, findings, conclusions, or recommendations expressed in this paper are those of the authors and do not necessarily reflect the views of the NVIDIA Academic Program.

## Data availability

Data will be made available on request.